\documentclass{article}

\usepackage{microtype}
\usepackage{graphicx}
\usepackage{subfigure}
\usepackage{booktabs} 
\usepackage{hhline, multirow}
\usepackage{adjustbox}

\usepackage{hyperref}

\usepackage[accepted]{icml2024}

\usepackage{amsmath}
\usepackage{amssymb}
\usepackage{mathtools}
\usepackage{amsthm}

\usepackage[capitalize,noabbrev]{cleveref}

\theoremstyle{plain}

\theoremstyle{definition}

\theoremstyle{remark}

\usepackage[textsize=tiny]{todonotes}

\icmltitlerunning{Expand-and-Cluster: Parameter Recovery of Neural Networks}

\begin{document}

\twocolumn[
    \icmltitle{Expand-and-Cluster:\\
    Parameter Recovery of Neural Networks}



\icmlsetsymbol{equal}{*}

\begin{icmlauthorlist}
    \icmlauthor{Flavio Martinelli}{epfl}
    \icmlauthor{Berfin \c{S}im\c{s}ek}{epfl,nyu}
    \icmlauthor{Wulfram Gerstner}{equal,epfl}
    \icmlauthor{Johanni Brea}{equal,epfl}
\end{icmlauthorlist}
    
\icmlaffiliation{epfl}{Department of Life Sciences and Computer Sciences, EPFL, Lausanne, Switzerland.}
\icmlaffiliation{nyu}{Center for Data Science, NYU, New York, United States}

\icmlcorrespondingauthor{Flavio Martinelli}{flavio.martinelli@epfl.ch}

\icmlkeywords{Machine Learning, ICML, Model Stealing Attacks, Teacher-Student Setup, Reverse Engineering, Non-Convex Optimization, Activation Function Symmetries, Symmetries, Overparameterisation, Parameter Recovery}

\vskip 0.3in
]



\printAffiliationsAndNotice{\icmlEqualContribution} 

\begin{abstract}
\looseness=-1
Can we identify the weights of a neural network by probing its input-output mapping?
At first glance, this problem seems to have many solutions because of permutation, overparameterisation and activation function symmetries. 
Yet, we show that the incoming weight vector of each neuron is identifiable up to sign or scaling, depending on the activation function.
Our novel method `Expand-and-Cluster’ can identify layer sizes and weights of a target network for all commonly used activation functions.
Expand-and-Cluster consists of two phases: 
(i) to relax the non-convex optimisation problem, we train multiple overparameterised student networks to best imitate the target function; 
(ii) to reverse engineer the target network's weights, we employ an ad-hoc clustering procedure that reveals the learnt weight vectors shared between students -- these correspond to the target weight vectors.
We demonstrate successful weights and size recovery of trained shallow and deep networks with less than 10\% overhead in the layer size and describe an `ease-of-identifiability' axis by analysing 150 synthetic problems of variable difficulty. 
\looseness=-1
\end{abstract}

\section{Introduction}
\label{sec:intro}

\looseness=-1
It is known since the 1980s that finding a solution to the XOR problem with gradient descent is easier with a larger hidden layer, 
even though a minimal network with two hidden neurons is theoretically sufficient to solve the problem \citep{rumelhart1985learning}.
Indeed, even very small networks have a non-convex loss function \citep{fukumizu2000local,mei2018landscape,frei2020agnostic,yehudai2020learning,simsek2024should}. 
Recent advances in the theory of artificial neural networks indicate that the loss function is rough -- i.e. predominantly populated by saddle points -- for networks of minimal size \citep{simsek2021geometry},
but becomes effectively convex in the limit of infinitely large hidden layers \citep{jacot2018neural, chizat2018global, du2018gradient, rotskoff2018trainability}. 
In a teacher-student setup, where teacher and student share the same architecture, the complexity of the loss landscape is linked to the ratio between the amount of permutation-induced critical points (zero gradients, non-zero loss) and the number of global minima at zero population loss \citep{simsek2021geometry}.
Importantly, as the width of the student increases, this ratio undergoes a substantial change: from much larger than one to very close to zero; suggesting that already for mild overparameterisation the amount of interconnected global minima largely dominates the amount of critical points \citep{cooper2018loss,simsek2021geometry}. 
This is consistent with our empirical observations: without overparameterisation, students trained to imitate the target (teacher) network get stuck in local high-loss minima.
Instead, if we expand the student width to four times the width of the target network we can reliably reach near-zero loss (Fig. \ref{fig:4}). 
At zero loss, target and student networks are functionally identical; but they are networks of different sizes and parameters.
Here we ask the following question: ``Is it possible to recover the weights and width of the target network from (near-zero) loss, overparameterised students?''
\looseness=-1

To answer this question, we first characterise the equivalence class of zero-loss overparameterised students -- i.e. all possible weight structures that can preserve functional equivalence of a given hidden layer. 
Three types of symmetries arise at the neuron level: 
\textit{(i) Permutation symmetries:} student neurons copying the weights of teacher neurons can be found in arbitrary order within the hidden layer; 
\textit{(ii) Overparameterisation symmetries:} redundant neuron groups can duplicate input weights to imitate a teacher neuron or cancel each others' contribution. 
\textit{(iii) Activation function symmetries:} even, odd or scaling symmetries in the activation function give rise to combinations of neurons that degenerate into constant or linear outputs, weight vectors of opposite signs with respect to teacher neurons or scaling factors transferred to adjacent layers; and combinations thereof.
As shown by \citet{simsek2021geometry} in a restricted setting, at zero loss each teacher neuron is duplicated by the student at least once. 
Importantly, we show that each teacher neuron input weight vector is preserved in zero-loss students up to a sign and/or scaling factor. 
Even on smoother overparameterised landscapes, exact zero loss is not numerically achievable in practice. Therefore, we developed a clustering procedure to separate neurons that are consistently found across $N$ different student networks from redundant units that do not persist across networks. The former are the ones copying the teacher's hidden neuron weights.

\begin{figure}[t]
    \begin{center}
    \centerline{\includegraphics[width=0.4\textwidth]{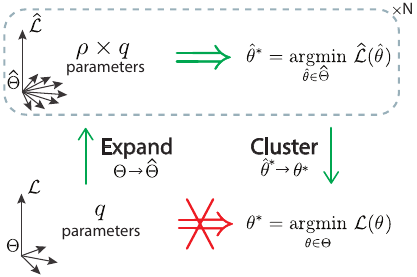}}
    \label{fig:1}
    \end{center}
    \vskip -0.25in
    \caption{{\bf Expand-and-Cluster}: we overcome the non-convex problem of recovering the $q$ parameters of an unknown network by: \textbf{(i)} expanding the dimensionality of the parameter space $\Theta$ by a factor $\rho$ to relax the optimisation problem, $\Theta \rightarrow \hat{\Theta}$; \textbf{(ii)} mapping the loss minimiser in expanded space $\hat{\theta}^*$ to the original parameter space via clustering of $N$ overparameterised solutions, $\hat{\theta}^* \rightarrow \theta^*$.}
\end{figure}

\looseness=-1
In summary, we tackle the highly non-convex parameter identification problem with a technique that resembles a popular strategy in the field of applied mathematics \citep{lovasz1991cones}: we {\bf expand}, or lift, the optimisation space to higher dimensions, find a solution through standard optimisation techniques, and {\bf cluster} to project the solution back to the original parameter space (Fig. \ref{fig:1}). 
Our contributions can be summarised as follows:
\vspace{-8pt}
\begin{itemize}
    \item A small, constant overparameterisation factor is enough to solve the non-convex problem due to the combinatorial proliferation of global minima, in agreement with \citet{simsek2021geometry};
    \item Building upon previous works \citep{petzka2020notes, simsek2021geometry}, we provide a complete formulation of overparameterisation and activation function symmetries. We discover a covert symmetry for activations functions composed of a linear and an even term (e.g. ReLU, GELU, softplus, SiLU);
    \item 
    We introduce Expand-and-Cluster, the first parameter recovery method suitable for any activation function to solve the identification problem in shallow, deep feed-forward and convolutional layers of unknown widths.\looseness=-1
\end{itemize}
  
\section{Motivation}
\label{sec:motivation}

{\bf 1. Understanding loss landscapes:}
In teacher-student setups, the ratio of global minima to other zero-gradient points is speculated to be an indicator of `trainability' of a given student -- i.e. to positively correlate to the probability of a training procedure to converge to a global minimum \citep{simsek2021geometry}. 
However, this quantification gives only a \textit{static} view of the trainability question, ignoring any effects that overparameterisation might induce to the \textit{dynamics} of training.
We empirically validate these speculations by showing that overparameterisation is a good indicator of student trainability for a rich family of differently shaped teachers (Fig. \ref{fig:4}). \looseness=-1

\vspace{-3pt}
{\bf 2. Neuroscience:}
Retrieving the connectivity of neural circuits is a daunting task.
Neuroscientists either infer the connectivity with massive connectomic imaging endeavours \citep{shapson2024petavoxel}, or try to estimate it from partial recordings of unit activities \citep{haspel2023reverse}. 
Although the second approach has many advantages, it is unclear how accurately one can reverse engineer the connectivity from unit activation recordings.
To start simple, we tackle this question in a deep-learning setting. In contrast to \citet{rolnick2020reverse}, we provide a learning-based solution adaptable to any activation function that relies on natural stimulation protocols, for example, the data the teacher is trained on. We see this work as a fundamental step towards reverse engineering brain circuits. \looseness=-1

\vspace{-3pt}
{\bf 3. Model stealing attacks:}
Testing the extent to which it is possible to steal information from deployed models has important practical consequences \citep{carlini2024stealing}. 
For example, once all parameters of a deployed network are identified, stronger and malicious adversarial attacks (requiring gradient or architecture information) can be performed on a deployed service \citep{chakraborty2018adversarial}, posing security threats and ethical issues. 
Given the small scale of reconstructed models in the field, this is not yet an issue of concern. \looseness=-1

\section{Related Work}
\label{sec:background:related}
{\bf1. Fundamental distinction with pruning and distillation:} 
Despite the apparent similarities, our model identification problem differs substantially from the classic pruning setting \citep{hoefler2021sparsity}. 
In pruning, one trades the number of parameters with a tolerated increase in test loss. 
In distillation settings, the focus is also on generalisation performance \citep{hinton2015distilling}. Moreover, students are often smaller and of a different architecture than the teacher \citep{beyer2022knowledge}. Even in the case of self-distillation \citep{furlanello2018born}, the student is not expected to be functionally equivalent to the teacher \citep{stanton2021does}.
In our setting, any increase in imitation loss or change in architecture is unacceptable, as it would result in reconstructing a potentially good approximation of the target network but not its parameterisation. 
Loss-unaware pruning schemes such as magnitude pruning \citep{lecun1989optimal, han2015learning,frankle2019lottery} or structural pruning based on weight properties \citep{srinivas2015data, mussay2021data} have no means of preserving the target parameters within the student. 
While loss-aware structural pruning techniques \citep{hu2016network, chen2022otov2} could have more control in trading pruning and loss increase, they come with no clear guarantee to remove all the non-trivial redundant structures described by the symmetries and, at the same time, preserve the target network parameters. 
Given the lack of alternative methods for non-relu activations, pruning constitutes one baseline for comparison in our experiments.

{\bf2. Non-convex optimisation and loss landscapes:} 
  Many non-convex optimisation problems are tackled with the following strategy: (i) expand, or \textit{lift}, to a higher dimensional space to relax the problem and guarantee convergence to global minima; (ii) map, or \textit{project}, the relaxed solution to the original space by exploiting the problem's intrinsic symmetries and geometry \citep{zhang2020symmetry}. This approach is used in applied mathematics \citep{lovasz1991cones, lasserre2001global}, machine learning \citep{ janzamin2015beating, zhang2020symmetry} and many others \citep{provablenonconvex}.
  Despite the above-mentioned achievements, for neural networks, the picture is far from complete.
  Unlike infinitely-wide neural networks \citep{jacot2018neural}, the loss functions of finite-width networks exhibit several non-convexities causing the gradient flow (from different initialisations) to converge to local minima non-identical to one another \citep{safran2018spurious, arjevani2021analytic, abbe2022initial}.  
  Yet, overparameterised solutions found by different initializations are similar in function space \citep{allen2020towards}, they can be \textit{approximately} mapped to each other by permutation of hidden neurons and exhibit the linear mode connectivity phenomenon \citep{singh2020model, wang2020federated, entezari2022role, ainsworth2022git, jordan2022repair}.    
  Even though the mildly overparameterised regime is non-convex, we show that we can exploit its reduced complexity to find zero-loss solutions in our identification setup; and that any zero-loss student can be \textit{exactly} mapped to every other zero-loss student. \looseness=-1

{\bf3. Interpretability:} 
  Explaining in qualitative terms the behaviour of single neurons embedded in deep networks is a challenging task \citep{olah2018building, zhang2021survey}. For example, in symbolic regression, small networks with vanishing training loss are desirable for interpretability \citep{udrescu2020ai, liu2024kan}. 
  Complementary to the notion of `superimposed-features' neurons found in underparameterised students \citep{elhage2022superposition, simsek2024should}, we provide explanations about the role of each hidden neuron found in zero-loss overparameterised students relative to a teacher network of minimal size. \looseness=-1
  
{\bf4. Functionally Equivalent Model Extraction:} 
  Our work focuses on functionally equivalent extractions, that is retrieving a model $\mathcal{M}$ such that $\forall x \in X, \mathcal{M}(x) = \mathcal{M}^*(x)$, where $\mathcal{M}^*(x)$ is the target model. This type of extraction is the hardest achievable goal in the field of Model Stealing Attacks \citep{oliynyk2022know}, using only input-output pairs \citep{jagielski2020high}. 
  Out of all the functionally equivalent models we aim to extract the one of \textit{minimal size}.
  Conditions for neural network identifiability and their symmetries have been studied theoretically for different activation functions \citep{sussmann1992uniqueness, fefferman1994reconstructing, zhong2017recovery, bui2020functional,petzka2020notes, vlavcic2021affine, vlavcic2022neural, bona2021parameter, stock2022embedding}, although overparameterised solutions were considered only in \citet{tian2019luck} and \citet{simsek2021geometry}. 
  Existing functionally equivalent extractions of trained networks rely on identifying boundaries between linear regions of shallow ReLU networks \citep{baum1991neural, jagielski2020high} and  single output deep ReLU networks \citep{rolnick2020reverse, carlini2020cryptanalytic}. 
  \citet{janzamin2015beating} show a theoretical reconstruction based on third-order derivatives.
  \citet{fornasier2022finite}, building upon \citep{fornasier2019robust,fu2020guaranteed,fornasier2021robust,fiedler2023stable}, propose an identification method for wide committee machines (shallow networks with unit norm second layer weights), where knowledge of the teacher layer size is necessary. 
  Anachronistically, \citet{tramer2016stealing} learn small, shallow teacher networks (20 hidden nodes) with overparameterised students but do not claim any parameter recovery. 
  Shallow networks with \textit{polynomial} activation functions can be globally optimised under some guarantees by lifting the optimisation problem to tensor decomposition \citep{janzamin2015beating, mondelli2019connection}. 
  We are the first to propose a recovery method for arbitrary activation functions on shallow and deep fully connected networks of unknown layer widths.
  Our method is learning-based and fundamentally different from the known approaches. 
  However, none of these methods were shown to work in large-scale applications.

\section{Symmetries of the identification problem}
\label{sec:symmetries}  

\begin{figure*}[t]
    \begin{center}
    \centerline{\includegraphics[width=\textwidth]{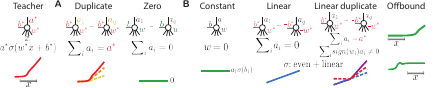}}
    \caption{\textbf{Catalogue of student neuron types at zero loss:} zero loss overparameterised students can only contain a few neuron types. Left: a teacher neuron is defined along with a sketch of its output, the grey bar indicates the finite input support $x$ to the neuron; the same colour-coded letters indicate equal quantities. \textbf{A) Neuron types from \citet{simsek2021geometry} adapted with biases:} \textit{duplicate-type} neurons combine to replicate a teacher neuron by copying its weight vector $w^*$ and bias $b^*$, their activations $a_i$ sum up to the teacher activation $a^*$. \textit{Zero types} have aligned weight vectors and biases but cancel each other via output weights. \textbf{B) Novel neuron types:} \textit{constant types} contribute a fixed amount to the next layer by learning a null vector. For even + linear activation functions, \textit{linear type} groups combine to contribute a linear function. \textit{Linear duplicate type} groups copy the teacher vector and its opposite, replicating the teacher neuron up to a linear mismatch. \textit{Offbound types:} at non-exact zero loss, their input support is placed in the linear or zero region of $\sigma$. 
    }
    \label{fig:2}
    \end{center}
    \vskip -0.2in        
\end{figure*}  

\begin{figure*}[t]
    \begin{center}
    \centerline{\includegraphics[width=0.8\textwidth]{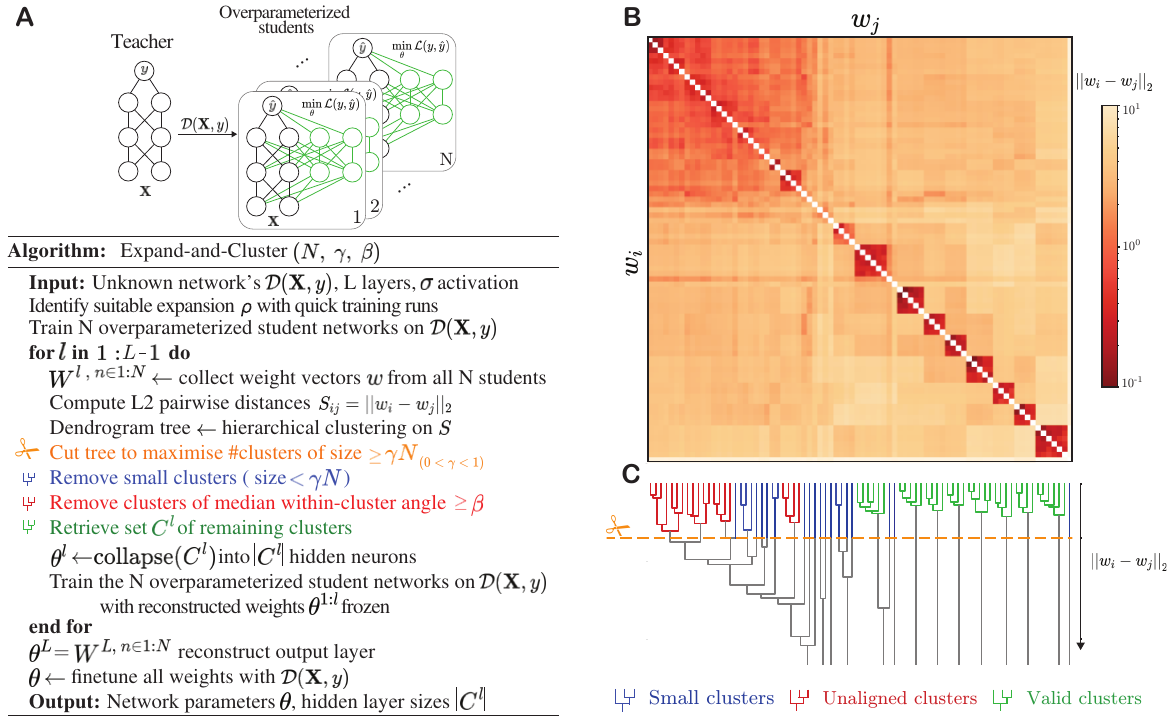}}
    \caption{\textbf{Parameter identification with Expand-and-Cluster.} \textbf{A) Training scheme:} once an overparameterisation factor yields near-zero training losses, train N overparameterised students on the teacher-generated dataset $\mathcal{D}(\mathbf{X}, y)$; \textbf{B) Similarity matrix}: L2-distance between hidden neurons' input weight vectors of layer $l$ for all $N$ students. Large-sized clusters are good candidate weight vectors. \textbf{C) Dendrogram obtained with hierarchical clustering:} the selected linkage threshold is shown in orange. Clusters are eliminated if too small (blue) or unaligned (red), the remaining clusters are shown in green. 
    The code is available at \url{https://github.com/flavio-martinelli/expand-and-cluster}.
    }
    \label{fig:3}
    \end{center}
    \vskip -0.2in        
\end{figure*}

To characterise the symmetries of the identification problem we start by reviewing the results of \citet{simsek2021geometry}, which are for neurons with no biases and asymmetric activation function. We will later extend this formulation to the practical deep learning setting and describe a convert symmetry that arises when the activation function is a combination of a linear and an even function.
For teacher-student setups, we call a student `overparameterised' if it has more hidden neurons than the teacher in at least one layer. 
If an overparameterised student network replicates the teacher mapping with zero loss, the space of all possible solutions is fully described by the geometry of the global minima manifold \citep{simsek2021geometry}. 
The global minima manifold contains only two types of hidden units, namely duplicate and zero-type neurons (see theorem 4.2 of \citet{simsek2021geometry} and Fig. \ref{fig:2}A); under the following assumptions:
one-hidden layer network $\sum_{i=1}^m a_i \sigma(w_ix)$, infinite input data support, population loss limit, no bias and analytical activation function $\sigma$ with infinite non-zero even \textit{and} odd derivatives evaluated at zero. 
The last assumption guarantees that the activation function has no symmetries around zero.
The intuition for the result of \citet{simsek2021geometry} is that, for zero-loss solutions, each teacher hidden neuron $a^*\sigma(w^*x)$ must be copied in the student by a \textit{duplicate-type} group of one or more units, contributing $\sum_i a_i\sigma(w_i x)$. 
The duplicates' input weight vectors are all aligned with the teacher neuron, $w_i=w^* \ \forall i$, while their weighted contribution equals the teacher neuron's output weight $\sum_i a_i = a^*$. 
In the same student there can also exist \textit{zero-type} neuron groups with a null contribution to the student input-output mapping, characterised by $w_1=\dots=w_q$, $\sum_i a_i =0$. 
These neuron types are summarized in Figure \ref{fig:2}A. 

\looseness=-1
We extend the categorization of neuron types of \citet{simsek2021geometry} to neurons with \textit{bias}, \textit{finite} input data support and activation functions that can be decomposed into \textit{even} or \textit{odd} functions plus a constant or a linear component.
This generalization includes commonly used activation functions such as ReLU, GELU, SiLU, sigmoid, tanh, softplus, and others (see \cref{app:activationfunctions} for details).
The catalogue of new neuron types is sketched in Figure \ref{fig:2}B.
In the case of a teacher neuron with bias $a^*\sigma(w^*x+b^*)$, a group of \textit{duplicate-type} $\sum_i a_i \sigma(w_ix + b_i)$ has input weight vectors and biases aligned to the teacher:
$w_i = w^*, \ b_i=b^* \ \forall i; \ \sum_i a_i =a^*$.
The new \textit{zero-type} group has aligned, but arbitrary, weights and biases $w_1=\cdots=w_q, \  b_1=\cdots=b_q, \sum_i a_i =0$.
With biases, \textit{constant-type} student neurons can also arise: 
they have vanishing input weights $w=0$ and contribute a constant amount of $a\sigma(b)$ to the next layer; to keep an exact mapping of the teacher (zero loss), this constant contribution must be cancelled out in the next-layer biases or with another constant-type neuron. \looseness=-1

\textbf{Even + linear activation function}: when the activation can be decomposed into an even and a linear function $\sigma(z) = \sigma_{lin}(z) + \sigma_{even}(z)$, for the sake of exposition we let $\sigma_{lin}(z)=z$, three phenomena arise: (i) Neurons can combine to contribute a linear + even function in non-trivial ways: $a_1\sigma(wx+b) + a_2\sigma(-wx-b) = (a_1-a_2)(wx+b) + (a_1 + a_2)\sigma_{even}(wx+b)$. 
This allows two or more student neurons to combine in groups of positively aligned ($+wx+b$) and negatively aligned ($-wx-b$) neurons. 
If in such a group $\sum_i a_i=0$, the neurons cancel the even component of the function and contribute a linear function: \textit{linear-type} group. 
(ii) Linear-type groups can effectively `flip' the sign of the input weight vector and bias of another neuron in the layer: $\sigma(wx+b) - 2(wx+b) = \sigma(-wx-b)$. 
A linear-type group can therefore correct the contribution of a neuron that learns the opposite teacher vector, or, more generally, correct the linear component of a group of positively and negatively aligned duplicate neurons; the latter named \textit{linear duplicate-type} group (Fig. \ref{fig:2}B). 
(iii) These linear-type groups can ultimately be combined into an affine operation: $\Theta x + \beta = \sum_k^K [(a^*_k \times w^*_k) x + a^*_k b^*_k]$, where $k$ is the index of duplicate neurons of opposite weight vector and bias, $\times$ is the outer vector product, $\Theta \in \mathbb{R}^{d_{out} \times d_{in}}$, $\beta \in \mathbb{R}^{d_{out}}$.
Notably: $\mathrm{rank}(\Theta) = min(K, d_{in}, d_{out})$.
See \cref{app:activationfunctionseven} for a more detailed explanation.
The even + linear symmetry is found in ReLU, LeakyReLU, GELU, Softplus, SiLU/Swish and other commonly used activation functions.
A concrete numerical example of different neurons found by a softplus student is shown in Figure \ref{fig:supp4}.
\textbf{Odd (+ constant) activation function}: when $\sigma(x) = c + \sigma_{odd}(x)$, student neurons can duplicate the teacher neuron with flipped signs: $a^*\sigma(w^*x+b) = - a^*\sigma(-w^*x-b) + c$. This is the case for tanh and sigmoid.
\textbf{Positive scaling}: for piece-wise linear activation functions such as ReLU and LeakyReLU, a positive scalar can be transferred between input and output weight vectors: $a^*\sigma(w^*x+b^*) = c a^* \sigma(\frac{1}{c}(w^*x + b^*))$, where $c>0$.
Finally, in near zero-loss solutions, there can be \textit{offbound-type} neurons: their hyperplane, spanned by $wx+b=0$, is placed outside of the input domain of the neuron. This results in the activation function being used in its asymptotic part (often constant or linear). 
The symmetries just described define the functional equivalence class between zero-loss overparameterised students. We show empirically, that the redundancy given by overparameterisation symmetries facilitates gradient descent in converging to a minimal loss. 
Given activation function symmetries, ReLU networks have the most amount (see \ref{app:listactfun}). Whether or not this fact plays a role in the ability to avoid the non-convexities of training is a matter of future research.
Most importantly, the only way an overparameterised student can reach zero loss is by representing all the neurons at least once \citep{simsek2021geometry}. 
By getting rid of all the redundant terms (zero, constant, linear and duplicate groups), teacher neurons can be identified up to permutation within the layer, a sign and/or a scaling factor. 
A numerical example of mapping a zero-loss overparameterised student into the teacher is shown in Figure \ref{fig:supp:stages}.\looseness=-1

\section{Expand-and-Cluster algorithm}
\label{sec:expandandcluster}

If a student can be trained to exact zero loss, which is possible in small setups, it is almost trivial to identify the different neuron types, including the teacher neurons (see two exact numerical examples in figs. \ref{fig:supp4}, \ref{fig:supp:stages}).
However, in larger setups, overparameterised networks are difficult to train to exact zero loss because of computational budgets and limited machine precision. 
Therefore we need ways to identify the teacher neurons from imperfectly trained students.
These students present approximate duplicates of teacher neurons as well as neurons of other types that point in arbitrary directions.
In a group of $N$ imperfectly trained students, neurons that approximate the teacher can be found consistently across students, while redundant units have arbitrarily different parameterisations.
Therefore we propose the following procedure (Fig. \ref{fig:3}):

{\bf Step 1: Expansion phase.}
Find the correct overparameterisation by rapidly training a sequence of networks with increasing sizes of hidden layers to a fixed convergence criterion. 
To do so, use teacher-generated input-output pairs (teacher queries), $\mathcal{D} = \{ \mathbf{X}, y \}$,
and minimize the mean square error loss between un-normalized outputs (e.g. before the softmax operation) of teacher and student networks.
This allows finding a network width $m$ at which convergence to nearly zero loss is possible (Fig. \ref{fig:supp1}B).
\looseness=-1

{\bf Step 2: Training phase.} 
Train $N$ students of $L$ layers, width $m$, on teacher queries $\mathcal{D} = \{ \mathbf{X}, y \}$ to minimize the loss to the lowest value achievable with classic optimiser techniques (Fig. \ref{fig:3}A). \looseness=-1

{\bf Step 3: Clustering phase.}
Collect the first unreconstructed hidden layer neurons of the $N$ students, then cluster the input weight vectors with hierarchical clustering on the L2 distance. 
In the case of even or odd symmetries, we want to avoid finding different clusters for neurons of merely opposite signs. 
Therefore we arbitrarily align all input weight vectors to have a positive $n^{\text{th}}$ element. 
This operation, which we call `canonicalization', maintains the functional equivalence, since there are ways to compensate for the sign flip for both odd and even symmetries (see section above). 
In the case of positive scaling symmetry, the clustering is performed on the cosine distance.
With a threshold selection criteria that maximizes the number of large clusters (size $\geq \gamma N,$  {\scriptsize{$0\negthinspace < \negthinspace\gamma \negthinspace \leq \negthinspace 1$}}), we obtain groups of aligned weight vectors; these clusters should include all duplicate teacher neurons. 
Proceed to filter out clusters whose elements are not aligned in angle (median alignment $\geq \beta$), removing eventual zero or constant type neuron clusters. 
Then, merge each remaining cluster of duplicate neurons into single hidden neurons by choosing the element in the cluster belonging to the lowest loss student.
Due to the unidentifiability of weight vector signs for even+linear symmetries, we can recover the network up to an affine transformation $\Theta x + \beta$. Therefore, to preserve the functional equivalence of the network, we solve the linear regression problem to find the affine transformation that minimises the error between the overparameterised students and the network with the newly reconstructed layer.
We noticed that higher layers align with the teacher weights only at prohibitively low losses \cite{tian2020student}. 
Therefore, if the student networks have more than one hidden layer left to reconstruct, we go back to Step 2 and train again $N$ overparameterised students of $L \leftarrow L-1$ layers, using as input $\mathbf{X}$ the output of the last reconstructed layer. 
Repeat this procedure until the last hidden layer is reconstructed (Fig. \ref{fig:3} Algorithm 1, more details in \cref{app:deepnetworksdetails}). 

{\bf Step 4: Fine-tuning phase.}
Adjust all the reconstructed network parameters using the training data. \looseness=-1

\section{Results} \label{sec:results}

\subsection{Synthetic teacher experiments}
\label{sec:artificialdata}

\begin{figure*}[t!]
    \begin{center}
    \centerline{\includegraphics[width=\textwidth]{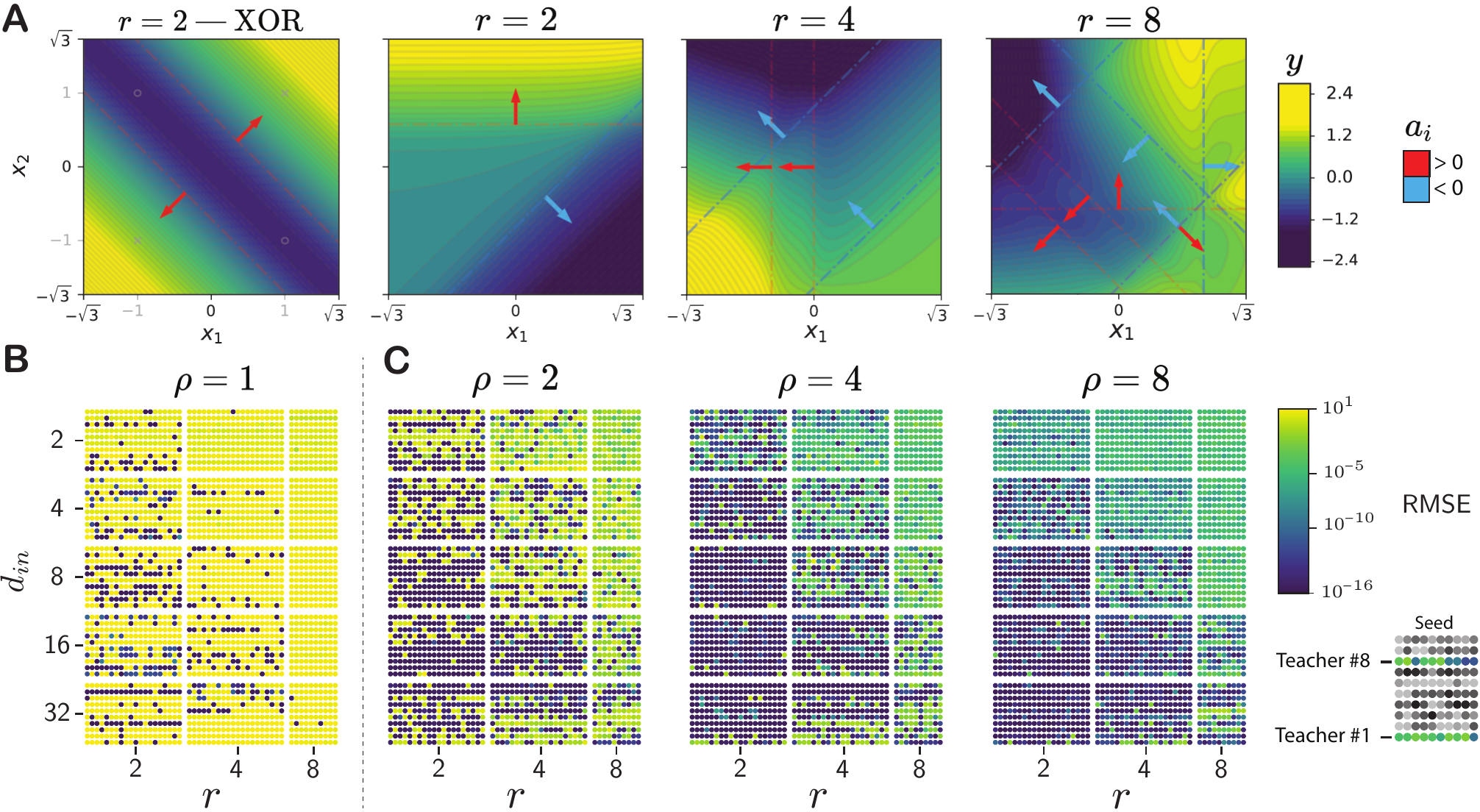}} 
    \caption{\textbf{Synthetic teachers define tasks of variable difficulty. A) For fixed $\mathbf{d_{in}}$, teacher complexity increases with number $\mathbf{r}$ of hidden neurons:} contourplot of the teacher network output. Each hidden neuron generates a hyperplane, ${w_i}^T x + b_i = 0$ (dashed lines); the direction of the weight vector $w_i$ is indicated by an arrow starting from the hyperplane and the sign of the output weight $a_i$ by its colour.  Top left: generalization of the XOR or parity-bit problem to a regression setting. From left to right: As the number of teacher hidden neurons $r$ increases the contour lines become more intricate. \textbf{B) Non-convexity prevents training to zero loss:} for each combination of $d_{in} =2,4,8,16,32$ and $r=2,4,8$ we generated 10 teachers; for each teacher, we trained 20 or 10 students (for $r=8$) with different seeds. Each teacher corresponds to one row of dots while each dot corresponds to one seed (see inset bottom right). Dark blue dots indicate loss below 10$^{-14}$. Student networks of the same size as the teacher ($\rho=1$) get often stuck in local minima. The effect is stronger for larger ratios $r/d_{in}$. \textbf{C) Effects of overparameterisation on convergence: }student networks with overparameterisation $\rho \geq 2$ are more likely to converge to near-zero loss than those without. We report the following general trends: (i) overparameterisation avoids high loss local minima, (ii) the dataset complexity, i.e. number of hidden neurons per input dimension $r / d_{in}$, determines the amount of overparameterisation needed for reliable convergence to near-zero loss. For difficult teachers, i.e. overcomplete ($r / d_{in} \geq 1$), training is very slow and convergence is not guaranteed in a reasonable amount of time (see Fig. \ref{fig:supp2}).
    }\label{fig:4}
    \end{center}
    \vskip -0.2in        
\end{figure*}

To test the effects of overparameterisation on the traversability of landscapes, we devised a series of very challenging regression tasks inspired by the parity-bit problem (or multidimensional XOR), known to be a difficult problem for neural networks \citep{rumelhart1985learning}. We construct synthetic one-hidden layer teacher networks of varying input dimension $d_{in}$, and hidden neuron number $r$. Our construction yields XOR-like and checkerboard-like functions where teacher neurons' hyperplanes are often parallel to each other and divide the input space into separate regions (Fig. \ref{fig:4}A), see \cref{app:artificial} for more details. 
In contrast to our approach, constructing shallow networks with randomly drawn input weight vectors yields easy tasks since all weights tend to be orthogonal \citep{saad1995line, goldt2019dynamics,raman2019fundamental}.
For this experiment we use the asymmetric activation function $\mathrm{g}(x) = \mathrm{softplus}(x) + \mathrm{sigmoid}(4x)$.

We trained overparameterised students on the family of teachers described above (Figs. \ref{fig:4} and \ref{fig:supp5}). For an overparameterisation factor $\rho$, the student hidden layer has $m = \rho r$ neurons.
To not venture away from the theoretical setting of near-zero loss, we trained all the networks with the ODE solver MLPGradientFlow.jl \citep{brea2023mlpgradientflow}. This allowed us to find global and local minima with machine precision accuracy for networks without overparameterisation (Fig. \ref{fig:4}B, $\rho = 1$). 
However, even with optimised solvers and slightly larger networks, it becomes challenging to converge fully to global minima within a reasonable amount of time (Fig. \ref{fig:4}C, $\rho \in \{2, 4, 8\}$). 
Hence, methods to deal with imperfectly trained students are needed. 
Since training is full-batch (30k data points), the only source of randomness is in the initialization; see \cref{app:trainingdetails} for more details. 
Figure \ref{fig:4}C shows a beneficial trend as overparameterisation increases, but also highlights a strong dependence on the dataset (or teacher) complexity $r/d_{in}$: 
as the number of hyperplanes per input dimension increases, it becomes harder for students to arrange into a global minimum configuration.

\begin{figure*}[t!]
    \begin{center}
    \centerline{\includegraphics[width=\textwidth]{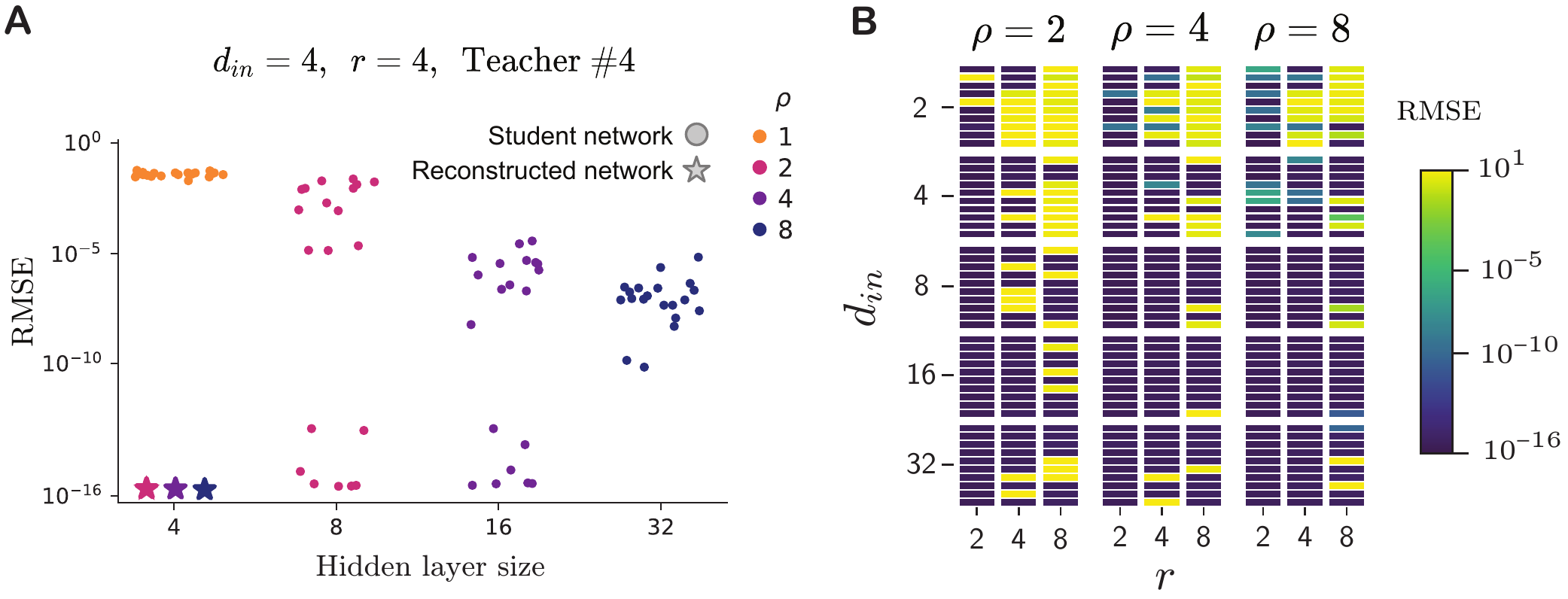}}
    \caption{{\bf A) Expand-and-Cluster applied to mildly overparameterised students reaches zero loss:} a total of 80 student networks with 4, 8, 16 or 32 hidden neurons have been trained using data generated by a teacher with $r=4$ hidden neurons and $d_{in}=4$ input dimensions. None of the 20 students with 4 hidden neurons reached zero loss (orange dots, $\rho=1$), while all overparameterised student networks have zero loss with 4 hidden neurons after reconstruction (large coloured stars). {\bf B) Loss after Expand-and-Cluster for all teacher networks and student sizes from Figure \ref{fig:4}:} the colour of each small horizontal bar represents the final loss. Only a small fraction of teacher networks (i.e., those in yellow) were not identified correctly.}
    \label{fig:5}
    \end{center}
    \vskip -0.2in        
\end{figure*}

We find that direct training of 20 student networks without overparameterisation (teacher with $r=4$ hidden neurons and input dimensionality $d_{in}=4$) does not yield a single case of convergence to zero loss (Fig. \ref{fig:5}A, $\rho=1$, hidden layer size $4$).
For the same teacher, Expand-and-Cluster can reconstruct the network up to machine-error zero loss and correct hidden-layer size if an overparameterisation of $\rho \geq 2$ is used in Step 2 of the algorithm (Fig. \ref{fig:5}A, star-shaped data-points).
This suggests that successful retrieval of all parameters of the teacher is possible (Fig. \ref{fig:5}A). 
We test the quality of parameter identification with Expand-and-Cluster for each teacher network of Figure \ref{fig:4} and illustrate the final loss of the reconstructed networks in Figure \ref{fig:5}B. 
For example, for $\rho=4,8$ and of the 30 teachers with input dimensionality $d_{in}=8$, all except 2 networks were correctly identified as indicated by a zero-loss solution (RMSE $\leq 10^{-14}$, dark blue in Fig. \ref{fig:5}B). 
Of 150 different teachers, 118 ($ \sim 80 \% $) were correctly identified with $\rho=4$.
In all but 7 out of 118 successful recoveries, the number of neurons found matches that of the teacher; the other cases have at most up to 4 neurons in excess (these can be easily categorised into zero-type and constant-type neurons, see e.g. Fig. \ref{fig:supp4}).
One can trade reconstruction loss vs. excess neurons by tuning the $\beta$ and $\gamma$ parameters of Expand-and-Cluster (not shown).
We conclude that, given a reasonably low loss, parameter identification is possible. 
Overparameterisation plays a key role in enabling gradient descent to find a global minimum, but this effect weakens as the ratio $r/d_{in}$ increases. 
We note that given more training time, even the hardest teachers could be learnt with overparameterisation $\rho=8$ (Fig. \ref{fig:supp2}). \looseness=-1

\subsection{Weight identification of trained networks}
\label{sec:mnist}

To show how the procedure scales to bigger applications, we recover parameters of networks trained on the MNIST \citep{lecun1998mnist}, FashionMNIST \citep{xiao2017fashion} and CIFAR10 \citep{krizhevsky2009learning} datasets. 
We pre-trained different one-hidden layer teachers composed of $r$ hidden neurons and different activation functions $\sigma$.
For parameter recovery, we must have access to the classifier's probabilistic output (e.g. the values before or after the softmax) and not only the most probable class (e.g. values after an argmax).
We then used input-output pairs generated by the last layer of the teacher to define a regression task for students of overparameterisation $\rho=4$.
After applying Expand-and-Cluster we obtain reconstructed networks of hidden layer size $\hat{m}$. 
We compare the reconstructed networks with teachers in Table \ref{tab:1}, and show low RMSE loss (mismatch between teacher logits and reconstructed network's logits is on the order of $10^{-4}$), close to perfect size recovery $\hat{m}/r \approx 1$ and low average cosine-distance $<\!d\!>$ between each teacher and reconstructed network neuron input-weight vectors. 
More metrics are shown in Table \ref{tab:3}.
Note that the difficulty of the training dataset the teacher is trained on is not crucial for the identification process, as the student networks are trained on  teacher-generated labels.
The slight reduction in average cosine alignment $<\!d\!>$ for CIFAR10 teachers is likely due to the higher dimensionality of the input images (3 channels instead of 1).
Experiments in Table \ref{tab:1} show successful identification when the input queries to the teacher are done with the same dataset the teacher was trained on.
Depending on the application, one may not have access to the teacher's training data.
We show in Table \ref{tab:2} that even if the teacher is trained on MNIST, the student can be trained on teacher queries from FashionMNIST and still achieve on par reconstruction performance.
We speculate that not any type of input query will be suitable to collect informative data-points from the teacher. 
Further research is needed to understand the quantity and quality of queries needed for successful identification. \looseness=-1

\begin{table}[h!]
  \vspace{-6pt}
  \caption{{\bf Identification of shallow networks:} successful reconstruction to low RMSE $\mathcal{L}$, similar size $\hat{m}/r \approx 1$ and low average cosine distance $d$ between identified and target neurons. All teachers are trained and queried with the same dataset, indicated in the leftmost column.}
  \vspace{1em}
  \begin{adjustbox}{width=\columnwidth,center}
  \begin{tabular}{cc|cccc}
  \label{tab:1}
  & $\sigma$ & $r$ & $\mathcal{L}$ & $\hat{m}/r$ & $<\!d(w_i, w^*_i)\!>$ \\
   \hhline{======}
  \parbox[t]{2mm}{\multirow{7}{*}{\rotatebox[origin=c]{90}{MNIST}}} 
  &$\mathrm{g}$        & $256$ & $1.5 \cdot 10^{-3}$        & $1.008$        & $2.1 \cdot 10^{-4}$ \\
  &$\mathrm{sigmoid}$  & $256$ & $8.4 \cdot 10^{-4}$        & $1.08$         & $3.8 \cdot 10^{-4}$ \\
  &$\mathrm{tanh}$     & $128$ & $1.7 \cdot 10^{-3}$        & $1.04$         & $1.3 \cdot 10^{-4}$ \\
  &$\mathrm{softplus}$ & $64$  & $2.3 \cdot 10^{-3}$        & $1.08$         & $1.4 \cdot 10^{-4}$ \\
  &$\mathrm{relu}$     & $64$  & $8.2 \cdot 10^{-3}$        & $1.12$         & $4.6 \cdot 10^{-4}$ \\
  &$\mathrm{gelu}$     & $64$  & $1.8 \cdot 10^{-3}$        & $1.08$         & $1.4 \cdot 10^{-4}$ \\
  &$\mathrm{leakyrelu}$& $64$  & $3.3 \cdot 10^{-3}$        & $1.01$         & $1.8 \cdot 10^{-4}$ \\
  \hhline{------}
  \parbox[t]{2mm}{\multirow{4}{*}{\rotatebox[origin=c]{90}{Fashion}}} 
  &$\mathrm{g}$        & $64$ & $4.7 \cdot 10^{-4}$        & $1.04$         & $3.3 \cdot 10^{-5}$ \\
  &$\mathrm{sigmoid}$  & $64$ & $2.5 \cdot 10^{-3}$        & $1.17$         & $6.8 \cdot 10^{-3}$ \\
  &$\mathrm{softplus}$ & $64$ & $3.7 \cdot 10^{-3}$        & $1.28$         & $1.5 \cdot 10^{-3}$ \\
  &$\mathrm{tanh}$     & $64$ & $3.3 \cdot 10^{-4}$        & $1.11$         & $4.8 \cdot 10^{-5}$ \\
  \hhline{------}
  \parbox[t]{2mm}{\multirow{4}{*}{\rotatebox[origin=c]{90}{CIFAR10}}} 
  &$\mathrm{g}$        & $64$ & $3.7 \cdot 10^{-3}$        & $1.00$         & $5.0 \cdot 10^{-3}$ \\
  &$\mathrm{sigmoid}$  & $64$ & $1.2 \cdot 10^{-3}$        & $1.06$         & $2.7 \cdot 10^{-3}$ \\
  &$\mathrm{softplus}$ & $64$ & $7.1 \cdot 10^{-3}$        & $1.06$         & $1.4 \cdot 10^{-2}$ \\
  &$\mathrm{relu}$     & $64$ & $2.3 \cdot 10^{-3}$        & $1.92$         & $2.4 \cdot 10^{-3}$ \\
  \end{tabular}
\end{adjustbox}
\vspace{-4pt}
\end{table}
\begin{table}[h!]
  \caption{{\bf Identification with a dataset different from the teacher training dataset:} a teacher network trained on MNIST is reconstructed by training students from teacher queries performed with a different dataset: FashionMNIST.}
  \vspace{1em}
  \begin{adjustbox}{width=\columnwidth,center}
  \begin{tabular}{c|cccc}
  \label{tab:2}
  $\sigma$ & $r$ & $\mathcal{L}$ & $\hat{m}/r$ & $<\!d(w_i, w^*_i)\!>$ \\
   \hhline{=====}
  $\mathrm{\;\;\;\;g\;\;\;\;}$        & $64$ & $1.14 \cdot 10^{-3}$        & $1.11$        & $2.26 \cdot 10^{-4}$ \\ 
  \end{tabular}
\end{adjustbox}
\end{table}

\begin{figure}[ht!]
  \begin{center}
  \centering
  \includegraphics[width=0.45\textwidth]{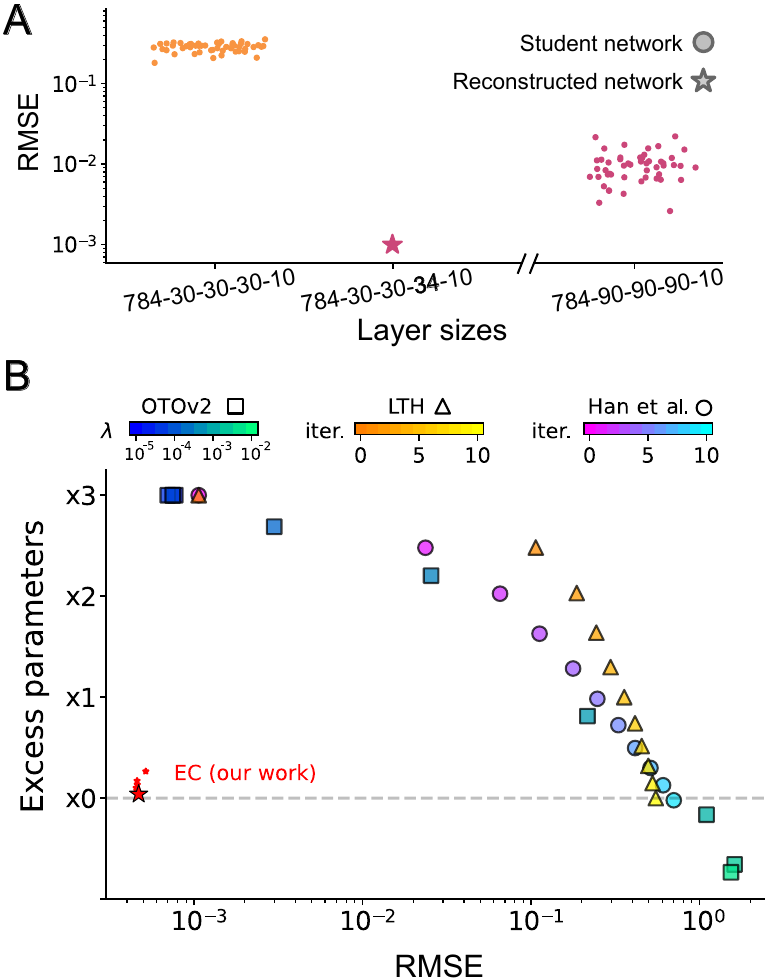}
  \caption{{\bf A) Identification of multiple hidden layers}: A deep teacher of layer sizes 784-30-30-30-10 is reconstructed with Expand-and-Cluster($N=50, \gamma=0.5, \beta=\pi/5$) applied to students of factor $\rho=3$ overparameterisation (magenta dots) with 4 excess neurons in the last hidden layer (magenta star). {\bf B) Baseline comparison with pruning techniques:} recovery of a shallow network trained on MNIST. Weight pruning methods: \citet{han2015learning}, lottery ticket hypothesis (LTH) \cite{frankle2019lottery} and structural pruning OTOv2 \citep{chen2022otov2} cannot achieve low imitation loss (RMSE) and equal teacher size (excess parameters = 0) simultaneously, while our algorithm Expand-and-Cluster (EC) is successful (star-points). Different colours indicate different hyperparameters of the methods (regularisation coefficient $\lambda$ for OTOv2 and pruning iteration number for the other two).\looseness=-1}
  \label{fig:6}
  \end{center}
  \vspace{-1em}
\end{figure}

Expand-and-Cluster can identify deep fully connected networks trained on synthetic data (see \cref{app:deepartificialrecovery}) and larger networks trained on MNIST, Figure \ref{fig:6}A (3 hidden layers of 30 neurons each, $\rho=3$, $\sigma = \mathrm{g}$). 
The reconstruction identifies layer sizes up to $4$ neurons in excess for the last hidden layer and achieves a loss of $3$ orders of magnitude lower than similar-sized networks trained from random initialization. 
Expand-and-Cluster can be applied ``as is" also to convolutional layers. Without loss of generality, one can treat each convolution channel as a hidden neuron of an MLP and the same symmetries described in Section \ref{sec:symmetries} apply. 
We show good preliminary reconstruction results for convolutional teacher layers in the appendix section \ref{app:conv}.
\looseness=-1

Given that our methodology for identification involves a compression step, we consider if currently available pruning methods are able to shrink overparameterised students back to the teacher size.
We compare Expand-and-Cluster to classic pruning techniques such as magnitude pruning \citep{han2015learning}, lottery ticket hypothesis \citep{frankle2019lottery} and the state-of-the-art structural pruning method OTOv2 \citep{chen2022otov2}. OTOv2 employs mixed $\ell_1/\ell_2$ regularisation at the level of units, combined with projected gradients to push units to zero.
To ensure a fair comparison in terms of computational budget, we compare one run of Expand-and-Cluster ($N=10, r=128, \sigma=\mathrm{tanh}$) with 10 runs of the other methods where we sweep their hyperparameters ($\rho=4$ for all students), Figure \ref{fig:6}B.
Expand-and-Cluster (EC) requires virtually no hyperparameter tuning, as $\gamma$ and $\beta$ are robust to changes in their values (Fig. \ref{fig:supp9}).
We also show that with $N \ge 5$ student networks Expand-and-Cluster identifies the teacher network with at most 10\% additional neurons (Fig. \ref{fig:supp8}). 
The comparison shown in Figure \ref{fig:6}B highlights a clear boundary between our method and current alternatives.
We speculate that the iterative nature of these pruning algorithms is not suited to the identification problem, as gradually shrinking the parameter space can lead to rougher landscapes that exhibit high non-convexity, leading to convergence to local minima. 
While Expand-and-Cluster fully exploits the expressivity of the overparameterised parameter space.
Notably, the weight pruning methods of \citet{han2015learning} and \citet{frankle2019lottery} are not capable of pruning entire units (Fig. \ref{fig:supp9}).

\section{Conclusions and future work}
\label{sec:conclusions}

Given data generated by a teacher of known size and architecture, it is usually impossible to recover its parameters by fitting a student network of the same size as the teacher with standard gradient-based training procedures. 
The detour to network expansion and clustering is our proposed approach to reliably find the teacher parameters.

Considering the symmetries involved, each neuron weight vector can be fully identified in networks with asymmetric activation functions, identified up to a sign for activation functions with odd or even+linear symmetries and identified up to a constant scaling for activation functions with positive scaling symmetry.
Convolutional layer symmetries are mappable to the ones listed in Section \ref{sec:symmetries}, by simply treating individual channel weight kernels as hidden neuron weight vectors.
Pooling and normalisation layers are not expected to introduce new symmetries.
Other symmetries induced by deep ReLU networks are discussed in \citet{grigsby2023hidden}, but those are mostly unidentifiable structures that do not contribute to the input-output mapping of a network.
To identify modern convolutional networks, an analysis of the symmetries of residual layers is needed. \looseness=-1

At this stage, our results are limited to small-scale setups, primarily due to the high computational budget required to reach low losses in bigger networks. 
A straightforward extension of this work is to focus on scaling to deeper networks and layer types, as well as speeding up the training process. 
Another future research direction concerns the query dataset needed to reconstruct the teacher network.
Theoretical developments are required to answer questions such as: what type of input queries are informative for the identification process? How can one generate more queries when access to teacher training data is limited or not granted? How many queries are sufficient?
Towards reverse engineering biological neural circuits, we show that we can solve the weight identifiability problem in artificial feed-forward networks.
However, further steps are needed to improve the reconstruction of connectivity from neural activity measurements, to cope, for example, with recurrent connectivity, unknown and noisy activations, partial observability or the integration of anatomical information. 
\looseness=-1
\clearpage

\section*{Acknowledgements}
This work was supported by Sinergia Project CRSII5\_198612 and SNF Project 200020\_207426.
The authors thank João Sacramento, Angelika Steger, Benjamin Grewe and Alexander Mathis for early feedback on the research. 

\section*{Impact Statement}
This paper presents work whose goal is to advance the field of Machine Learning. 
There are many potential societal consequences of our work, none which we feel must be specifically highlighted here.

\bibliography{main}
\bibliographystyle{icml2024}

\newpage
\appendix
\onecolumn
\section{Appendix}
\label{sec:appendix}

\counterwithin{figure}{section}
\counterwithin{table}{section}
\setcounter{figure}{0}

\subsection{Code availability}
The code is available at \url{https://github.com/flavio-martinelli/expand-and-cluster}.

\subsection{Activation functions} \label{app:activationfunctions}
\subsubsection{Even plus linear activation functions} \label{app:activationfunctionseven}

Considering a neuron with an activation function that can be decomposed into linear plus even terms: 
$$a\sigma(wx+b) = a c_1 \cdot (wx+b) + a \sigma_{\text{even}}(wx+b)$$ where $c_1$ is the slope of the linear approximation around $0$. 
The total contribution of aligned, $+(wx+b)$, and opposite, $-(wx+b)$, student neurons is:
\begin{align*}
  \sum_{i \in N^+} a_i \sigma(wx + b) + \sum_{j \in N^-} a_j \sigma(-(wx + b)) =&
  \underbrace {c_1\left(\sum_{i \in N^+} a_i - \sum_{j \in N^-} a_j\right)(wx + b)}_\textrm{Linear} \ \ \ \ + \\ 
  & \underbrace{ \sum_{k \in {N^{\pm}}} a_k \ \sigma_{\text{even}}(wx + b) }_\textrm{Even}
\end{align*}
where $N^+, N^-, N^\pm$ contain aligned, opposite or all neuron indices, respectively.

If $\sum\nolimits_{k \in {N^{\pm}}} a_k=0 $, we obtain a \textit{linear-type} group that contributes a linear function to the next layer; to keep an exact mapping of the teacher, there must be another \textit{linear-type} group in the same layer contributing the exact opposite linear term. 
If $\sum\nolimits_{k \in {N^{\pm}}} a_k= a^*$, then the neuron group is a \textit{linear duplicate-type}, replicating a teacher neuron up to a misaligned linear contribution; that can be accounted for by another linear group in the same layer. 
An example of different neurons reached by a $\mathrm{softplus}$ student is shown in supplementary Figure \ref{fig:supp4}.

\subsubsection{Constant plus odd activation functions} \label{app:activationfunctionsodd}
Considering a neuron with an activation function that can be decomposed into a constant plus odd term: 
$$a\sigma(wx+b) = a c_0 + a \sigma_{\text{odd}}(wx+b)$$ where $c_0 = \sigma(0)$. The total contribution of aligned and opposite student neurons is:
\begin{equation*}
   \sum_{i \in N^+} a_i \sigma(wx + b) + \sum_{j \in N^-} a_j \sigma(-(wx + b))= 
    \underbrace {c_0 \sum_{k \in {N^{\pm}}} a_k }_\textrm{Constant} \ \ + \ \  
   \underbrace{ \left(\sum_{i \in N^+} a_i - \sum_{j \in N^-} a_j\right) \sigma_{\text{odd}}(wx + b) }_\textrm{Odd}
\end{equation*}
where $N^+, N^-, N^\pm$ contain aligned, opposite or all neuron indices, respectively.

If $\sum\nolimits_{i \in {N^+}} a_i - \sum\nolimits_{j \in {N^-}} a_j=0$, we obtain a \textit{constant-type} group; to keep an exact mapping of the teacher, the sum of all \textit{constant-type} groups in the same layer and the next layer bias must add up to contribute the original bias of the next layer neuron. 
If $\sum\nolimits_{i \in {N^+}} a_i - \sum\nolimits_{j \in {N^-}} a_j=a^*$, then the neuron group is a \textit{duplicate-type}, replicating a teacher neuron up to a misaligned constant contribution; that can be accounted for by other constant groups in the same layer or biases in the next. 
For $\mathrm{tanh}$, there is no constant contribution term ($c_0=0$) while it is the case for $\mathrm{sigmoid}$.

\subsubsection{Classification of activation functions based on symmetries}
\label{app:listactfun}

\begin{itemize}
    \item \textbf{No symmetries} $\mathrm{sigmoid + softplus}$, $\mathrm{Mish}$ \citep{misra2019mish}, $\mathrm{SELU}$ \citep{klambauer2017self},
    \item \textbf{Odd}: $\mathrm{tanh}$ 
    \begin{itemize}
        \item[$\triangleright$] \textbf{Odd + constant}: $\mathrm{sigmoid}$
    \end{itemize}
    \item \textbf{Even + linear}: $\mathrm{GELU}$ \citep{hendrycks2016gaussian}, $\mathrm{Swish / SiLU}$ \citep{ramachandran2017searching} 
    \begin{itemize}
        \item[$\triangleright$] \textbf{Even + linear + constant}: $\mathrm{softplus}$
        \item[$\triangleright$] \textbf{Even + linear + positive scaling}: $\mathrm{ReLU}$, $\mathrm{LeakyReLU}$
    \end{itemize}
\end{itemize}

\subsection{Expand-and-Cluster detailed procedure}\label{app:expandandcluster}
The input weight vectors of all neurons in a given layer of $N$ mildly overparameterised students are clustered with hierarchical clustering \citep{murtagh2012algorithms} using the average linkage function $\ell (\mathcal{A}, \mathcal{B}) = \frac{1}{|\mathcal{A}| \cdot |\mathcal{B}|} \sum_{i \in \mathcal{A}}  \sum_{j \in \mathcal{B}} S_{ij}$, where $S_{ij} = \| w_i - w_j \|_2$ and $w_i$ and $w_j$ are input weight vectors of neurons in the same layer but potentially from different students (Fig. \ref{fig:3}B).
To identify the clusters of teacher neurons we look for the appropriate height $h$ to cut the dendrogram resulting from hierarchical clustering (Fig. \ref{fig:3}C).
With $K(h) = \{\kappa_1, \kappa_2, \ldots\}$ the set of clusters at threshold $h$, and $|\kappa_i|$ the size of cluster $i$, we define the set $C(h, \gamma N) = \{\kappa_i\in K(h)\; |\; |\kappa_i|\geq \gamma N\}$ of clusters larger than a fraction $\gamma \in (0, 1]$ of the number of students $N$.
For $\gamma > 1/N$, the set $C(h, \gamma N)$ usually does not contain small clusters of zero or offbound type neurons, because they are not aligned between different students.
We cut the dendrogram at the height that maximises the number of big clusters: $\hslash = {\arg \max}_h |C(h, \gamma N)|$.
The set $C(\hslash, \gamma N)$ may still contain clusters of non-aligned neurons whose input weights are close in Euclidean distance (e.g. approximate constant-type neurons with $\|w_i\|$ near zero) but not aligned in angle. Therefore we remove clusters whose within-cluster median angle is higher than $\beta$.
Each remaining cluster is then collapsed into a single hidden neuron with a winner-take-all policy: we choose the weight vector(s) that belong to the best-performing student in the cluster. Note that if multiple output weights $a_1,\dots,a_q$ of the same network belong to the same cluster, they can be combined into a single output weight $a = \sum_i a_i$, following the definition of duplicate neurons (Fig. \ref{fig:2}).
If the best student duplicates a given teacher neuron more than once, we take their average.
Alternatively, we have also tried to average over all neurons in each cluster and obtained a similar final performance. 
The hyperparameters of Expand-and-Cluster ($N, \gamma, \beta$) used for each experiment are summarised in \cref{app:expandandclusterhyperparams}. 

The biases could approximately be reconstructed by identifying all the constant-type neurons, but we found this procedure somewhat brittle.
Instead, we fine-tune all bias parameters (keeping constant the weight vectors) with a few steps of gradient descent on the reconstructed network. 
After each layer reconstruction, we retrain $N$ students with fixed non-trainable reconstructed layers but learnable biases, and the remaining layers are overparameterised and learnable. 
In this retraining phase, we monitor the learned bias values $b_k^n$ of the last reconstructed layer, if the same neuron $k$ learns different bias values across different students $n$ ($\text{std}_n(b_k^n)>0.1$) we consider that neuron badly clustered and therefore remove it from the layer. We repeat this procedure until the last hidden layer, the final output layer can be reconstructed by simply retrieving the output weights of the last hidden layer neurons. 
Ultimately further fine-tuning can be performed on the whole network to minimise the loss as much as possible. 

\subsection{Synthetic data in teacher-student networks}
\label{app:artificial}

All tasks with synthetic data have $d$-dimensional uniformly distributed  input data in the range $x_i \in \lbrack -\sqrt{3},  \sqrt{3} \rbrack $. 
A specific task is defined by the parameters of a teacher network. 
Each hidden neuron $i$ of the teacher is randomly sampled from a set of input weights $w_i \in \{-1, 0, 1\}^{d_{in}}$, output weights $a_i \in \{-1, 1\}$ and biases $b_i \in \{-\frac{2}{3}\sqrt{3}, -\frac{1}{3}\sqrt{3}, 0, \frac{1}{3}\sqrt{3}, \frac{2}{3}\sqrt{3}\}$. 
We repeat the sampling if two hidden neurons are identical up to output weights signs to avoid two hidden neurons cancelling each other. 
The resulting input weight vectors $w$ are first normalised to unity and then both $w$ and $b$ are multiplied by a factor of 3. 
The above procedure yields hyperplanes in direction $w$  located at a distance $|b|/||w||$ from the origin, and a steeply rising (or falling) activation on the positive side of the hyperplane. 
Finally, analogous to batch normalization, the output weights and biases are scaled such that the output has zero-mean and unit variance when averaged over the input distribution: $a \leftarrow a / \mathrm{std}(y)$ and $b_2 = - \langle y \rangle /\mathrm{std}(y)$, where $y$ is the output vector of the network.
We study teachers with input dimensionality $d_{in} \in \{2, 4, 8, 16, 32 \} $ and hidden layer size $r \in \{ 2, 4, 8 \}$. 
Figure \ref{fig:4}A shows examples of different teachers with input dimension $d_{in}=2$ and a single hidden layer. 
We use the asymmetric activation function $\sigma = \sigma_{sig}(4x) + \sigma_{soft}(x)$, where $\sigma_{sig} = \frac{1}{1+e^{-x}}$ and $\sigma_{soft} = log(1+e^x)$ for all our simulations unless specified otherwise. 
The construction of deep teacher networks follows the same mechanism as described above, with the additional extra step of scaling the weights such that each unit of the successive layer receives standardised inputs, analogous to what batch-norm would do for a full batch. 
This procedure ensures that the input support to each layer does not drift to values that are off-bounds with respect to where each hidden neuron hyperplane is placed. 
Classically trained networks also follow this scheme to have useful activations, but it is crucial to avoid potential support drift in arbitrarily constructed deep networks.

\subsection{Recovery of deep artificial teacher networks} \label{app:deepartificialrecovery}

\begin{figure*}[h]
  \begin{center}
  \centerline{\includegraphics[width=0.5\textwidth]{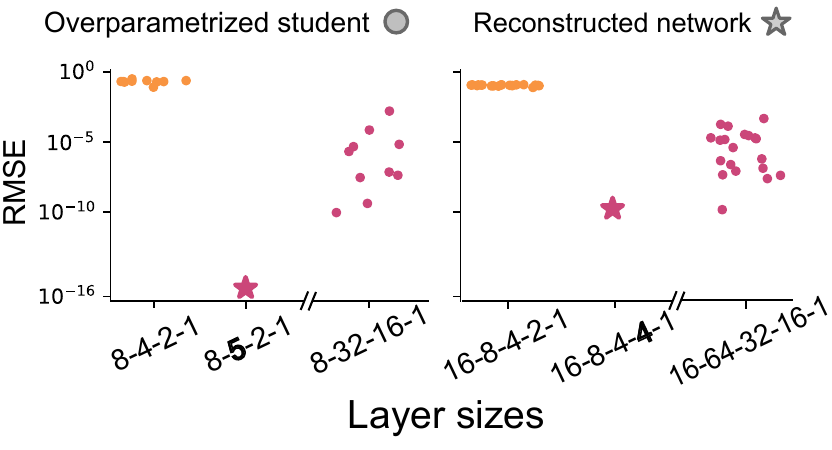}}
  \caption{{\bf Deep artificial teachers}: a teacher with input dimension 8, a first hidden layer with 4 neurons, and a second hidden layer with 2 neurons, and a single output (denoted as 8-4-2-1, left) was constructed by stacking the procedure of Figure \ref{fig:4} and generated data to train students of overparameterisation $\rho=8$ (denoted as 8-32-16-1). Similarly, a teacher with architecture 16-8-4-2 (right) generated data to train students with architecture 16-64-32-16. Expand-and-Cluster applied to students trained with an overparameterisation of $\rho=8$ identified the teacher network with one additional neuron in the first hidden layer (left) or with two additional neurons in the third hidden layer (right). In both cases, the loss is below $10^{-10}$ whereas direct training with the `correct' network architecture never reached a loss below $10^{-2}$.}
  \label{fig:supp6}
  \end{center}
  \vskip -0.2in       
\end{figure*}

We built two and three hidden layers networks by stacking the procedure to generate artificial teachers detailed in \cref{app:artificial}. Expand-and-Cluster \textit{without retraining step at each layer reconstruction} was applied to these networks: the reconstructed networks have one or two superfluous neurons for the two and three-layer teachers, respectively; (Fig. \ref{fig:supp6}). 
We emphasise that the final loss of the pruned student networks is at least 8 orders of magnitudes lower than the loss of the best out of 10 students trained with the {\em correct} number of neurons in each of the hidden layers. 
Thus, running 10 training runs with student networks overparameterised by a factor of $\rho =8$ enables us to correctly identify or nearly identify the teacher network, while direct training was not successful with the same number of runs (Fig. \ref{fig:supp6}). Detailed explanations on training and reconstruction are explained in the following sections (\cref{app:deepnetworksdetails} and \cref{app:expandandclusterhyperparams}).

\subsection{Training and reconstruction details} \label{app:trainingdetails}

\subsubsection{Synthetic teacher tasks}
To explore overparameterised networks trained to exact zero-loss or up to machine precision (for Float64 machine precision is at $10^{16}$), we integrated the gradient flow differential equation $\dot{\theta}(t) = -\nabla \mathcal{L}(\theta(t), \mathcal{D})$ with ODE solvers, where $\mathcal{L}$ is the mean square error loss. 
Specifically, we used the package MLPGradientFlow.jl \citep{brea2023mlpgradientflow} to follow the gradient with high accuracy and exact or approximate second-order methods to fine-tune convergence to a local or global minimum. 
All of the toy model networks are trained with Float64 precision on CPU machines (Intel Xeon Gold 6132 on Linux machines).

The networks of Figure \ref{fig:4} were trained on an input dataset $\mathbf{X}$ of 30k data points drawn from a uniform distribution between $-\sqrt{3}$ and $+\sqrt{3}$ and targets $y$ computed by the teacher network on the same input $\mathbf{X}$. 
Students were initialised following the Glorot normal distribution, mean 0 and std = $ \sqrt{\frac{2}{\text{fan\_in}+\text{fan\_out}}} $ \citep{glorot2010understanding}. 
We allocated a fixed amount of iteration steps per student: 5000 steps of the ODE solver \texttt{KenCarp58} for all networks, plus an additional 5000 steps of exact second order method \texttt{NewtonTrustRegion} for non-overparameterised networks ($\rho=1$) or 250 steps of \texttt{BFGS} for overparameterised networks ($\rho \geq 2$). 
The stopping criteria for the second training phase were: mean square error loss $\leq 10^{-31}$ or gradient norm $\nabla \mathcal{L}(\theta(t)) \leq 10^{-16}$. 
Each iteration step is full-batch, the only source of randomness in a student network is its initialization.

The networks shown in Figure \ref{fig:5}, after reconstruction, were fine-tuned for a maximum of 15 minutes with \texttt{NewtonTrustRegion} if the number of parameters was below or equal to 32 or \texttt{LD\_SLSQP} otherwise.

\subsubsection{Synthetic deep networks}\label{app:deepnetworksdetails}
Toy deep students were trained on a dataset generated in the same way as their shallow counterparts. 
We allocated 3 hours to solve the gradient flow ODE with \texttt{Runge-Kutta4} followed by 6 to 12 hours of approximate second-order optimisation with \texttt{BFGS}. 
After reconstruction, we gave 60 minutes of time budget to fine-tune with \texttt{LD\_SLSQP} (Fig. \ref{fig:6}A).

\subsubsection{Training of teachers and students}\label{app:mnisttraining}
Several teachers with a single hidden layer of $10, 30$ or $60$ neurons were trained with the activation function defined in \cref{app:activationfunctions} to best classify the MNIST dataset by minimizing the cross-entropy loss with the Adam optimiser \citep{kingma2015adam} and early stopping criterion. 
For each hidden-layer size, the best-performing teacher was used to train the students, see Figure \ref{fig:supp2}A. The student networks were trained to replicate the teacher logits (activation values before being passed to the softmax function) by minimizing a mean square error loss. 
Students were initialised with the Glorot uniform distribution, ranging from $-a$ to $a$, where $a= \sqrt{\frac{6}{\text{fan\_in}+\text{fan\_out}}}$. 
The training was performed with the Adam optimiser on mini-batches of size 640 with an adaptive learning rate scheduler that reduces the learning rate after more than 100 epochs of non-decreasing training loss. 
A maximum of 25k epochs was allocated to train these students on GPU machines (NVIDIA Tesla V100 32G). For the layerwise reconstruction of deep teacher networks, every retraining was given a maximum of 10k epochs. 

The reconstructed networks shown in Figure \ref{fig:6}B are fine-tuned with the same optimising recipe for a maximum of 2000 epochs. For deep reconstructed networks, a final finetuning of 15k epochs was performed on all the parameters.

\subsubsection{Expand-and-Cluster hyperparameters}\label{app:expandandclusterhyperparams}

The Expand-and-Cluster procedure has only 3 hyperparameters to tune: the number of student networks $N$, the cutoff threshold to eliminate small clusters $\gamma$ and the maximally admitted alignment angle $\beta$. 

\begin{figure}[h]
   \begin{center}
   \centerline{\includegraphics[width=0.5\textwidth]{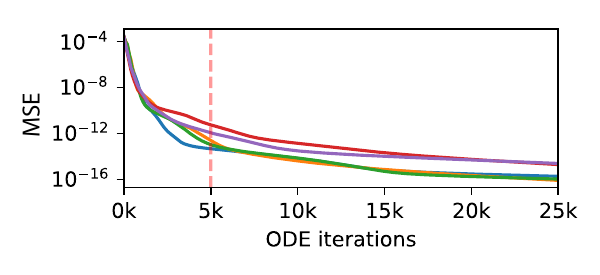}}
   \caption{{\bf Training longer slowly decreases student loss for difficult teachers:} the results shown in Figure \ref{fig:4} for difficult teachers are not at convergence despite the large amount of ODE solver steps dedicated (5K for every simulation). If training is continued we can see a slow decrease in loss, hinting that a very complex landscape is generated by difficult teachers. This simulation took more than a day to compute. Each colour corresponds to a different student loss curve, the teacher used was of $d_{in}=2, r=8$.}
   \label{fig:supp2}
   \end{center}
   \vskip -0.2in        
\end{figure}

\begin{itemize}
  \item \textbf{Shallow synthetic teachers:} all the procedure was performed with $N=10$ (for $r=8$) or $N=20$ (for $r=2, 4$), $\gamma = 0.8$ and $\beta = \pi / 24$. 
  \item \textbf{Deep synthetic teachers:} we reconstructed the synthetic deep networks \textit{without the retraining step at each layer reconstruction}. This required looser conditions, especially for the higher layers: for the synthetic deep networks of 2 hidden layers: $N=10$, $\gamma=0.4$ and $\beta=1-\arccos(0.01) \simeq 8^{\circ}$; for the 3 hidden layer case: $N=20$, $\gamma=1/3$, $\beta\simeq 0.26^\circ$ for the first layer and $\beta\simeq 30^\circ$ for the other layers. 
  The alignment gets quickly lost as depth increases, motivating the looser $\beta$ angles. We found this procedure too brittle and decided to modify the algorithm into the version in the main paper (Fig. \ref{fig:3}) by adding the retraining step at each layer reconstruction. The reconstruction results for deep MNIST teachers follow this updated algorithm. 
  \item \textbf{Shallow MNIST teachers:} for MNIST student networks we notice that no alignment of weight vectors can be expected for the corner pixels of the images because these pixels have the same value for nearly all input images. 
  This prevents weights connected to corner pixels from moving from the random initialization.
  Therefore we removed the uninformative pixels using the tree-based feature importance method Boruta \citep{kursa2010feature} to identify with statistical significance the informative features; the resulting map is shown in Figure  \ref{fig:supp2}B. 
  The parameters used for Expand-and-Cluster on MNIST shallow networks of Figure \ref{fig:6}B are $\gamma=0.5$ and $\beta=\pi / 6$ while $N$ sweeps from 2 to 20. 
  \item \textbf{Deep MNIST teacher:} all the procedure was performed with $N=50$, $\gamma = 0.5$ and $\beta = \pi / 5$ for every layer. Layerwise reconstruction alternated to the training of new students is a much more stable procedure than the one described for deep synthetic teachers as there is no need to tune the Expand-and-Cluster parameters for every layer.
\end{itemize}

\subsection{Convolutional layers reconstruction}
\label{app:conv}
\begin{figure}[]
    \centering
    \includegraphics[width=0.45\textwidth]{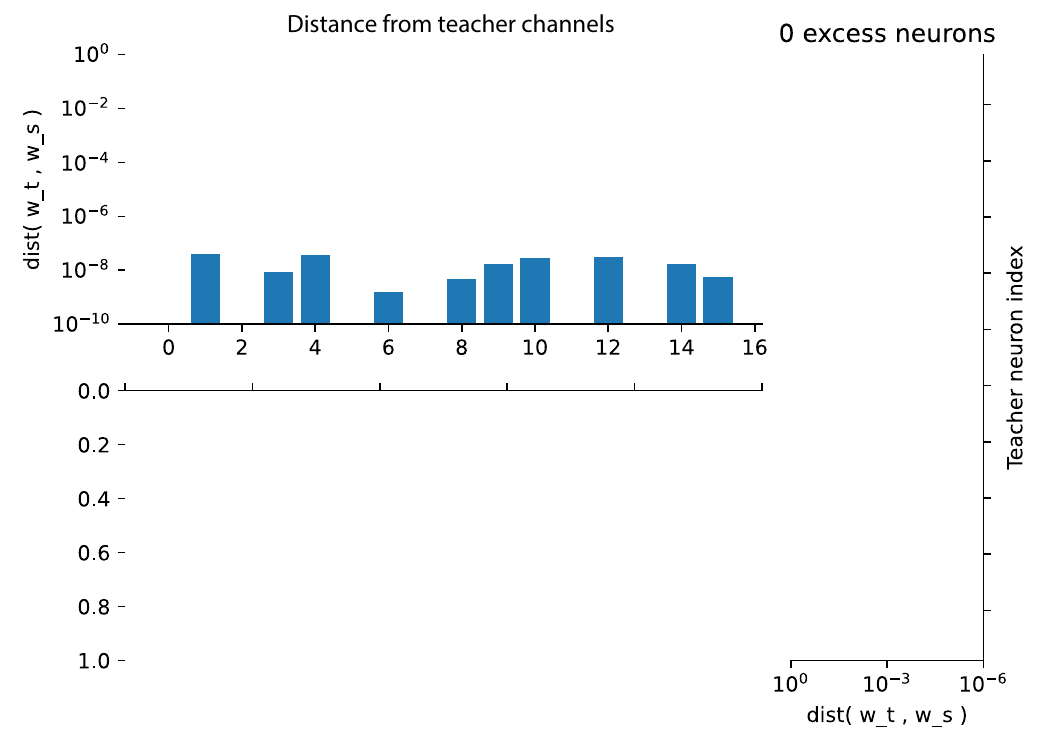}
    \hfill
    \includegraphics[width=0.45\textwidth]{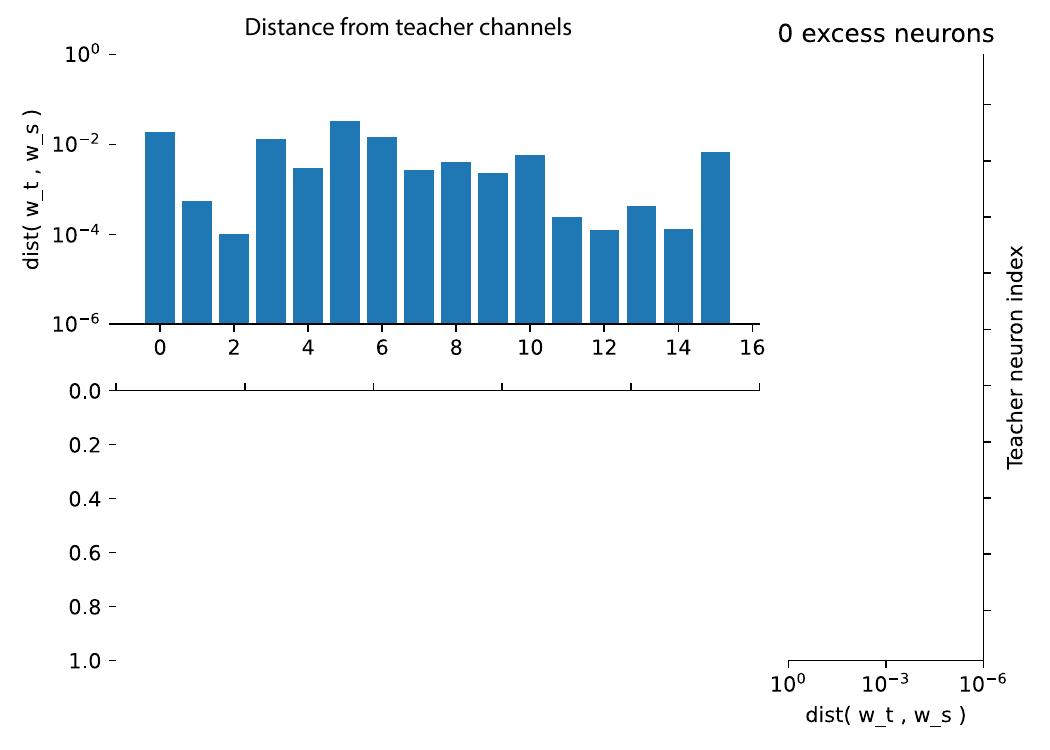} 
    \caption{{\bf Alignment of channel weights between teacher and reconstructed student}:  The figure shows the alignment of the different channel weights between the teacher and the reconstructed student for the convolutional network trained on the CIFAR10 dataset.  The left plot is conv-layer 1, and the right plot is conv-layer 2.}
    \label{app:convfig}
\end{figure}

We trained a convolutional neural network on the CIFAR10 dataset. The network is composed of two convolutional layers of 16 channels, each followed by a maxpool operation. After the convolutions, a fully connected layer of 32 hidden neurons computes the outputs.
Students were trained with overparameterisation factor $\rho=3$, we adapted the clustering algorithm to consider channels in the convolutional layers as analogous to hidden neurons of an MLP. The alignment of the different channel weights between the teacher and the reconstructed student is shown in Fig. \ref{app:convfig}.

\subsection{Supplementary figures and table}\label{app:supplementaryfigures}

\begin{figure*}[h]
   \begin{center}
   \centerline{\includegraphics[width=0.9\textwidth]{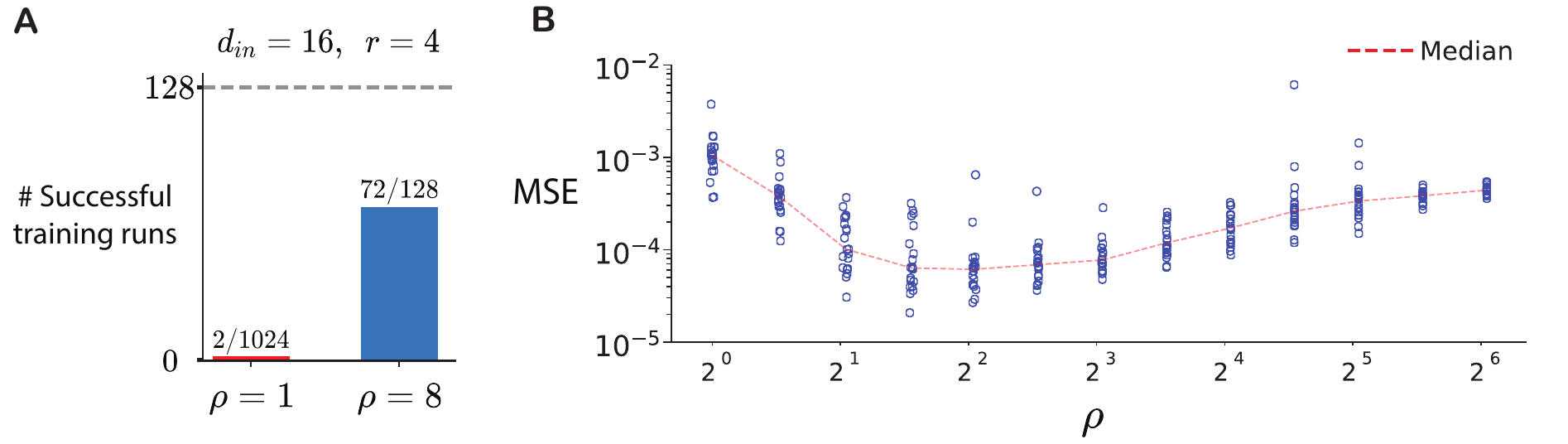}}
   \caption{{\bf A) overparameterisation performance scales beyond simple linear parameter scaling:} we successfully retrieve an identical input-output mapping of the original network with a student of the same size ($\rho=1$) only 2 times out of 1024 runs. If we instead train 8 times fewer networks ($n=128$) that are 8 times bigger ($\rho=8$) we achieve a significant improvement in success rate, 72 out of 128 runs. Note that the computational budget of the two experiments is equal. Example shown with teacher configuration $d_{in}=16, r=4, \text{teacher } \# 7$. {\bf B) Probing expansion factors:} It is impossible to know the overparameterisation factor of a student if the teacher is unknown. Nevertheless, one can probe different student sizes with quick training runs of stochastic gradient descent to identify a suitable student size that gives minimal losses. With a fixed time budget per training run, we notice an increase in MSE for large overparameterisation.}
   \label{fig:supp1}
   \end{center}
   \vskip -0.2in        
\end{figure*}

\begin{figure*}[h]
   \begin{center}
   \centerline{\includegraphics[width=0.75\textwidth]{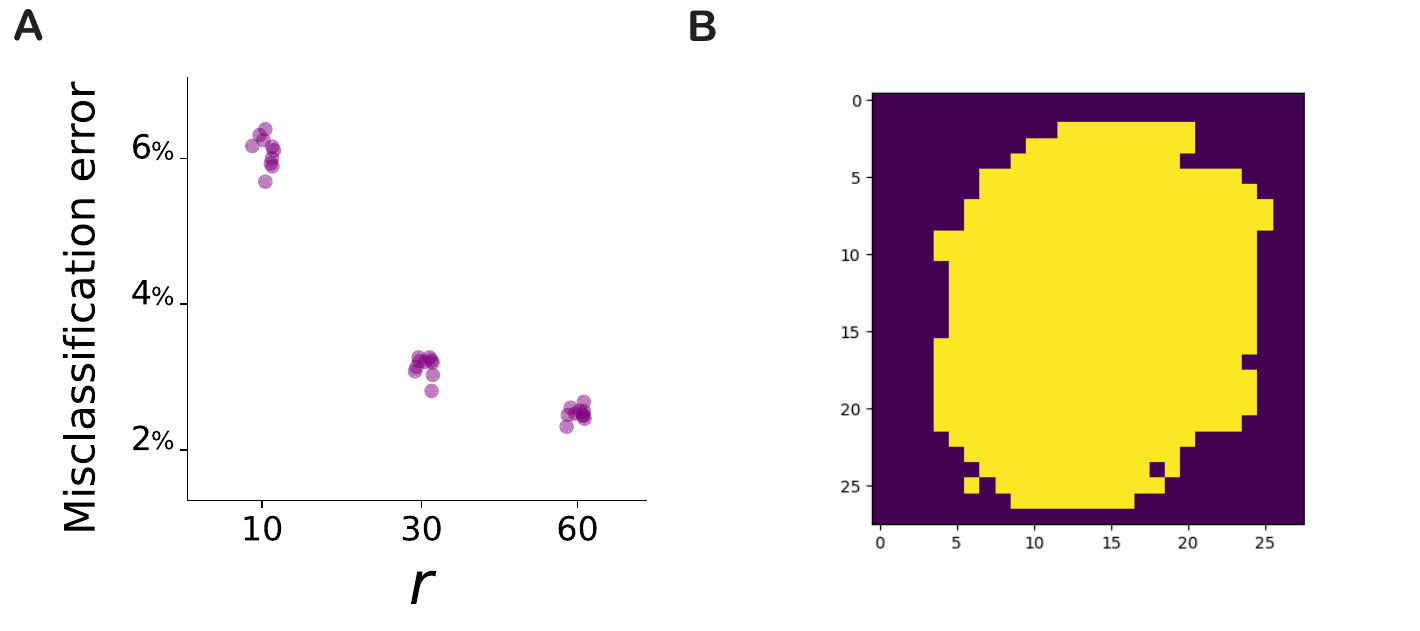}}
   \caption{{\bf A) MNIST teachers misclassification error on test set}. {\bf B) MNIST important pixels mask obtained with BORUTA \citep{kursa2010feature}:} only connections projecting from yellow pixels are considered for the Expand-and-Cluster procedure.}
   \label{fig:supp3}
   \end{center}
   \vskip -0.2in       
\end{figure*}

\begin{figure*}[h]
   \begin{center}
   \centerline{\includegraphics[width=\textwidth]{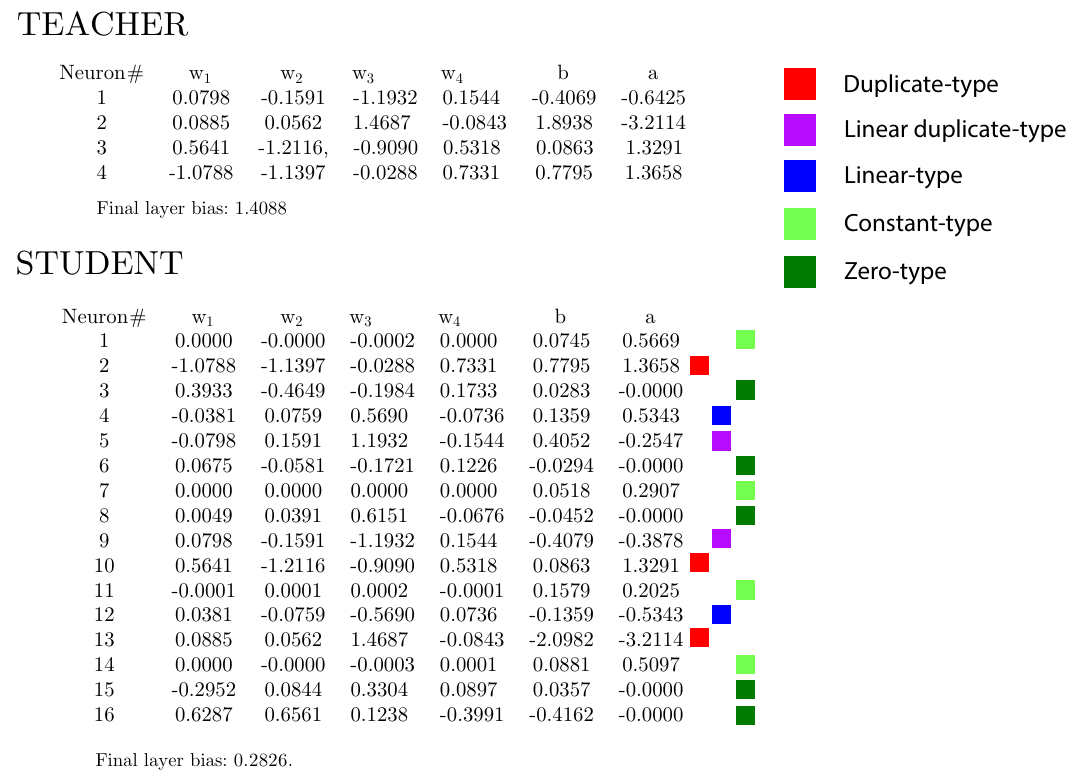}}
   \caption{{\bf Example parameters of a softplus teacher and student:} the table above shows the parameters of a $d_{in}=4, r=4$ $\mathbf{\mathrm{softplus}}$ teacher. 
   The table below shows the parameters of a student of $\rho = 4$ trained on the defined teacher to RMSE $\sim 10^{-11}$. 
   At a near-zero loss, we can classify all the different neurons (colour-coded) composing the student, following the classification scheme of Figure \ref{fig:2}.
   Note that the presence of a linear duplicate type implies the presence of a linear type, this group of 4 neurons (blue and purple coded) collaborates to replicate teacher neuron \#1.}
   \label{fig:supp4}
   \end{center}
   \vskip -0.2in        
\end{figure*}

\begin{figure*}[h]
  \begin{center}
  \centerline{\includegraphics[width=0.85\textwidth]{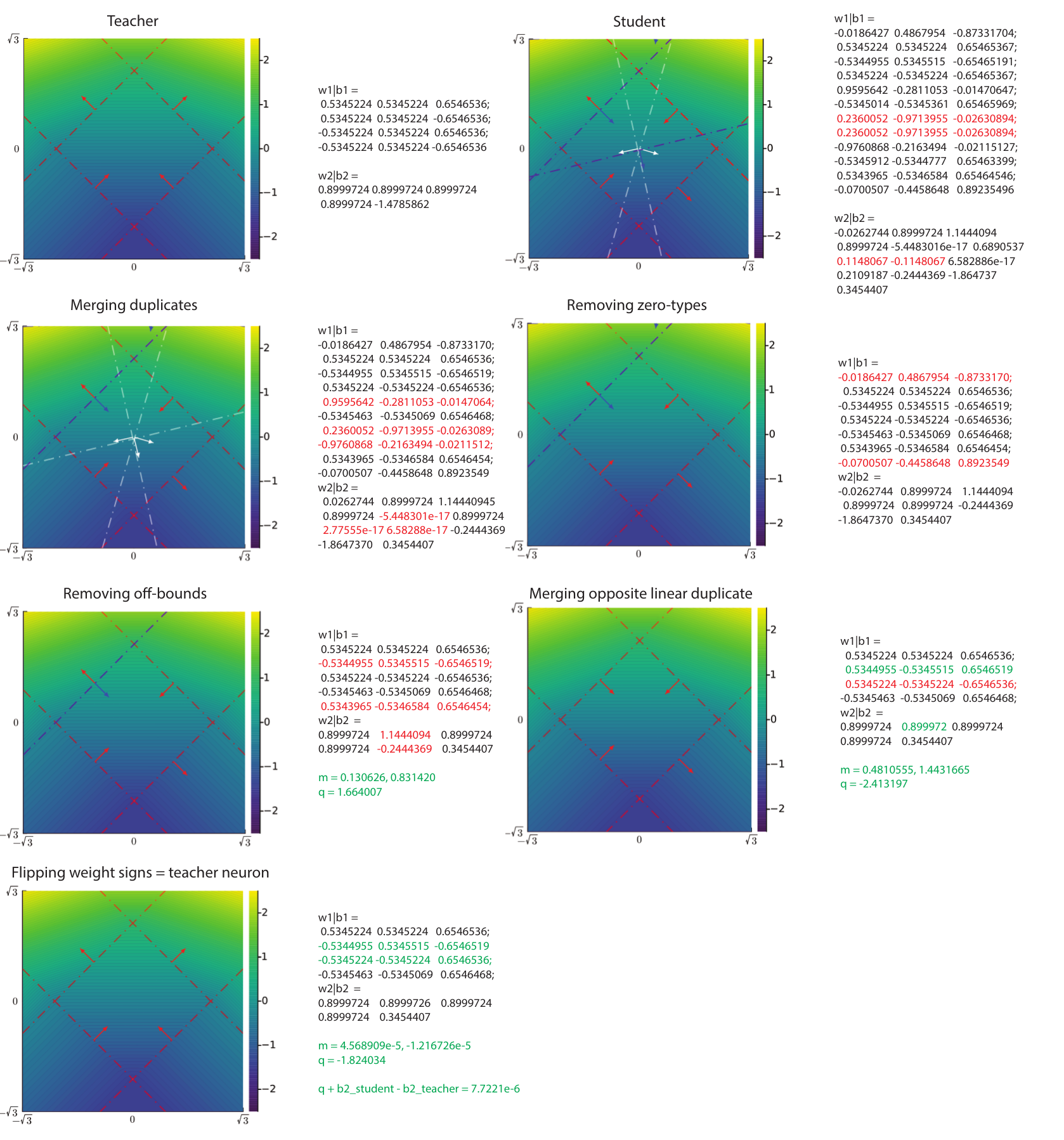}}
  \caption{
    {\bf Mapping an exact zero loss overparameterised relu student to the teacher using symmetries:}
    A numerical example of a $\sigma=\mathrm{relu}, r=4, d_{in}=2, d_{out}=1$ teacher. We follow the same network representation as in figure \ref{fig:4}: we plot each hidden neuron hyperplane ($wx+b=0$) and the input weight vector as an arrow, the colour of the arrow indicates if the scalar output weight is $>0$ (red), $<0$ (blue) or $\approx 0$ (white). 
    {\bf Top left}: representation of the teacher, each input and output weight vector (concatenated with bias) are shown numerically on the right side of each figure. 
    {\bf Top right}: $\sigma=\mathrm{relu}, m=12, d_{in}=2, d_{out}=1, \mathcal{L}=10^{-16}$ overparameterised student. 
    {\bf Following plots from left to right, top to bottom:} each plot shows a step towards mapping the student network to the teacher. Each step maintains functional equivalence by making use of the symmetries of Section \ref{sec:symmetries}. Red parameters indicate the parameters modified in the next step, green parameters are the parameters that have been modified with the current step (if both green and red should be present, we leave it in red). In order: \textit{Merging duplicates:} a pair of duplicate neurons is merged by summation of their output weights. \textit{Removing zero-types:} neurons with near-zero output weights (white arrows) are removed. \textit{Removing off-bounds:} off-bound neurons (one is barely visible at the top of the contour plot) are removed by taking care of their contribution. In this case, the off-bounds relu neurons have their hyperplane outside of the input domain and their weight vector pointing towards the input domain, hence they contribute a linear function to the output. Therefore a linear transformation $mx+q$ is added to maintain functional equivalence. \textit{Merging opposite linear duplicates:} Neurons that share a hyperplane but features pointing in opposite directions can be merged into a unique neuron + linear component. \textit{Flipping weight signs:} finally, the weight vectors opposite to the teacher are flipped by adding a linear component. We verify that the final linear component sums up to zero to confirm the functional equivalence of the mapping. 
  }
  
  \label{fig:supp:stages}
  \end{center}
  \vskip -0.2in        
\end{figure*}

\begin{figure}[t]
    \begin{center}
    \centering
    \includegraphics[width=\textwidth]{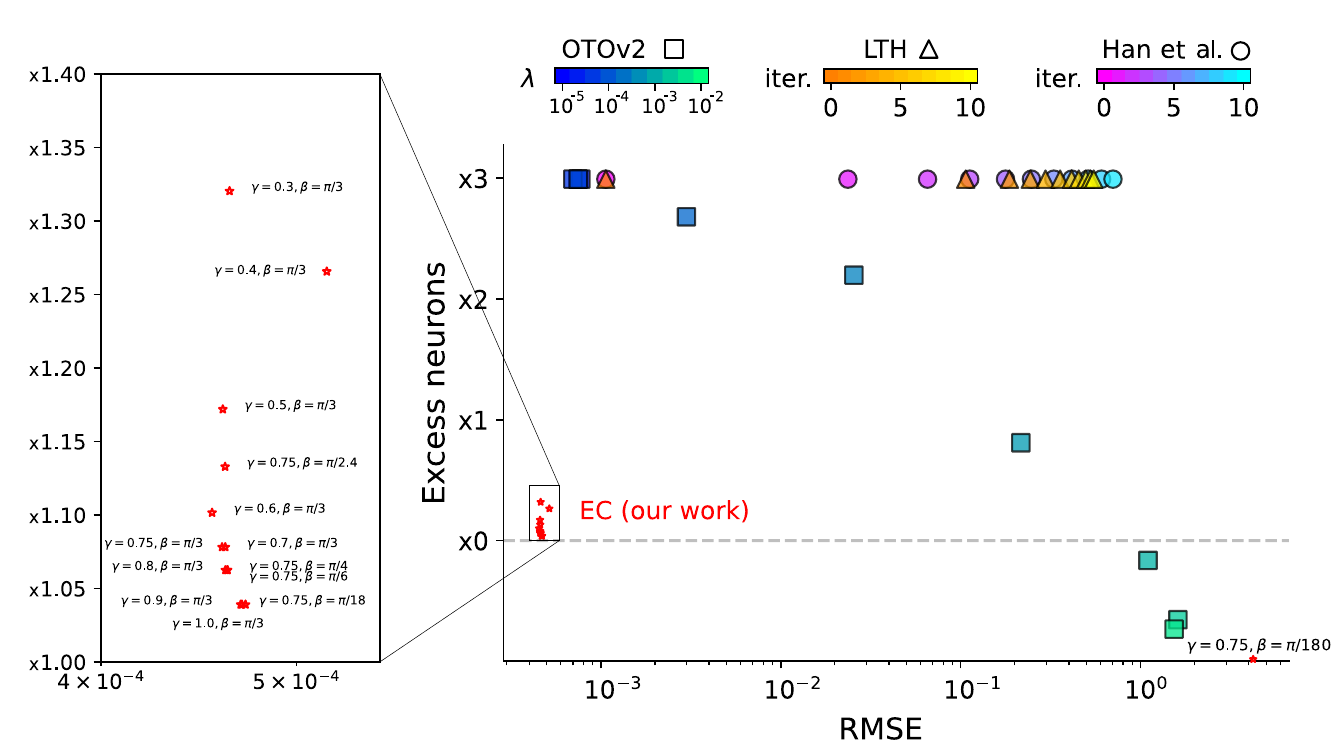}
    \caption{{\bf Neurons in excess with respect to other pruning methods:} same experiment as Figure \ref{fig:6}A but looking at the number of neurons in excess. The weight pruning methods lottery ticket hypothesis (LTH) \cite{frankle2019lottery} and classic magnitude pruning \cite{han2015learning} never effectively push all the weights of any unit to zero (as they were not designed to do so). The structural pruning method OTOv2 \cite{chen2022otov2} instead prunes only units. Training $N=10$ overparameterised students for Expand-and-Cluster (EC) is computationally equivalent to training $10$ runs of parameter search for the other methods. While EC and OTOv2 runs can be parallelised, for LTH and Han et al. they need to be run sequentially. Once $N=10$ overparameterised students are trained, testing different hyperparameter sets for EC is not as computationally expensive. Different runs of EC for different parameters $\gamma$ and $\beta$ show how EC hyperparameter search is not as crucial as a search for a regularisation coefficient $\lambda$. {\bf Inset -- Expand-and-Cluster robustness to hyperparameter sweep:} The star-shaped points highlight the stable robustness in the performance of EC for different hyperparameter settings. We sweep $\gamma \in [0.3, 0.4, 0.5, 0.6, 0.7, 0.8, 0.9, 1.0]$ while keeping $\beta=\pi/3$ and $\beta \in [\pi/2.4, \pi/3, \pi/4, \pi/6, \pi/18, \pi/180]$ while keeping $\gamma=0.75$. The only set of parameters that fails the reconstruction is $\gamma=0.75, \beta=\pi/180$. Looking at the role of these parameters in detail: $\beta$ filters out unaligned clusters which tend to have elements orthogonal to each other. Any value of $\beta \geq \pi/6$ is sufficient to filter out unaligned clusters without filtering potentially important duplicates. While $\gamma$ is involved in filtering out small clusters that we expect to be made of zero-type neurons, therefore any value of $\gamma$ that is too low ($\gamma \leq 0.4$) would risk filtering out important clusters, while a value of $\gamma$ too high risks to miss some clusters of duplicates.}
    \label{fig:supp9}
    \end{center}
    \vspace{-3mm}
  \end{figure}

  \begin{table}[H]
    \caption{{\bf Detailed statistics of experiments of Table \ref{fig:6}A}: $\mathcal{L}$ is the root mean square error loss of the student network, $\hat{m}/r$ is the number of neurons of the student divided by the teacher size, $<\!d(w_i, w^*_i)\!>$ and $\max_i d(w_i, w^*_i)$ are the average and maximum cosine distance between reconstructed and teacher input weight vectors (absolute value cosine distance in case of symmetries where the sign of the weight vector cannot be recovered), $<\!d(a_i, a^*_i)\!>$ and $\max_i d(a_i, a^*_i)$ are the same metrics for the output weights.}
    \vspace{1em}
    \begin{adjustbox}{width=\textwidth,center}
    \begin{tabular}{c|cccccccccc}
    \label{tab:3}
    $\sigma$ & $r$ & $N$ & $\gamma$ & $\beta$ & $\mathcal{L}$ & $\hat{m}/r$ & $<\!d(w_i, w^*_i)\!>$ & $\max_i d(w_i, w^*_i)$ & $<\!d(a_i, a^*_i)\!>$ & $\max_i d(a_i, a^*_i)$ \\ \hline
    $\mathrm{g}$        & $256$ & 20 & 0.5 & $\pi/3$ & $1.53 \cdot 10^{-3}$  & $1.008$  & $2.13 \cdot 10^{-4}$ & $1.92 \cdot 10^{-3}$ & $3.48 \cdot 10^{-8}$ & $2.41 \cdot 10^{-6}$                \\
    $\mathrm{sigmoid}$  & $256$ & 20 & 0.9 & $\pi/3$ & $8.37 \cdot 10^{-4}$  & $1.12$   & $3.77 \cdot 10^{-4}$ & $6.00 \cdot 10^{-3}$ & $5.49 \cdot 10^{-4}$ & $2.84 \cdot 10^{-2}$               \\
    $\mathrm{tanh}$     & $128$ & 10 & 0.9 & $\pi/3$ & $1.73 \cdot 10^{-3}$  & $1.04$   & $1.29 \cdot 10^{-4}$ & $9.26 \cdot 10^{-4}$ & $3.06 \cdot 10^{-5}$ & $3.91 \cdot 10^{-3}$                \\
    $\mathrm{softplus}$ & $64$  & 20 & 0.9 & $\pi/3$ & $2.97 \cdot 10^{-3}$  & $1.09$   & $1.08 \cdot 10^{-2}$ & $6.82 \cdot 10^{-1}$ & $5.66 \cdot 10^{-3}$ & $3.62 \cdot 10^{-1}$                \\
    $\mathrm{relu}$     & $64$  & 20 & 0.5 & $\pi/3$ & $8.16 \cdot 10^{-3}$  & $1.12$   & $4.63 \cdot 10^{-4}$ & $8.02 \cdot 10^{-3}$ & $3.34 \cdot 10^{-5}$ & $2.03 \cdot 10^{-3}$              
    \end{tabular}
    \end{adjustbox}
\end{table}

\begin{figure*}[h]
  \begin{center}
  \centerline{\includegraphics[width=0.5\textwidth]{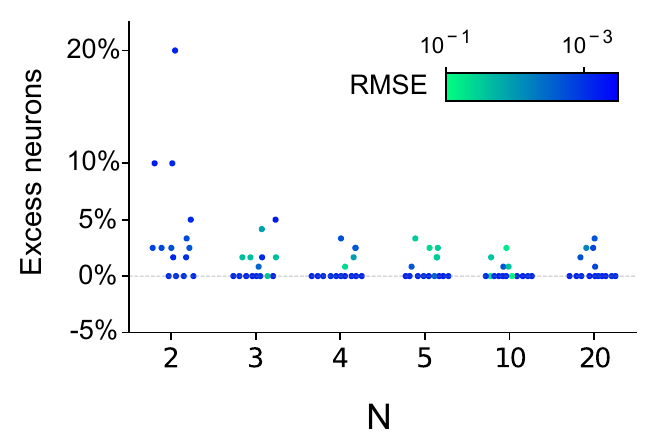}}
  \caption{{\bf A) MNIST shallow teachers}: fraction of excess neurons with respect to the teacher size) clustered as a function of the $N$ students used for Expand-and-Cluster($N, \gamma=0.5, \beta=\pi/6$). Combined statistics across three shallow teachers of sizes $r \in \{10,30,60\}$ pre-trained on MNIST data.}
  \label{fig:supp8}
  \end{center}
  \vskip -0.2in 
\end{figure*} 

\begin{figure}[H]
  \centering
  \includegraphics[width=\textwidth]{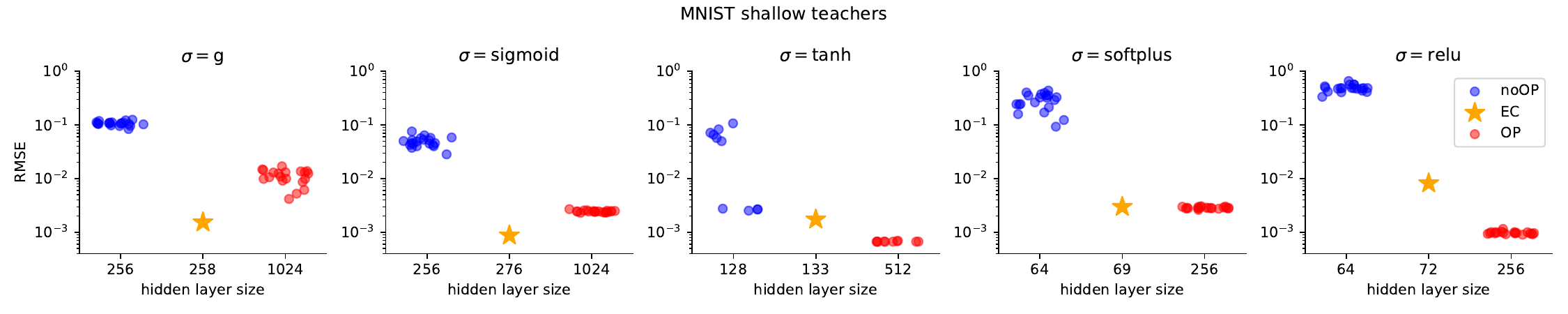}
  \caption{{\bf Loss and size comparisons for MNIST experiments of Table \ref{tab:1} of the main paper}: without overparameterisation (noOP, blue dots) training gets stuck at high loss values due to the extreme non-convexity of the landscape. These networks show poor imitation of the teachers and no alignment in weights. Only $\sigma=tanh$ seems to be an exception to this phenomenon. Overparameterisation is {\bf necessary} to reliably reach lower losses due to the proliferation of global minima in the landscape (OP, red dots). Expand-and-Cluster can successfully reconstruct the teacher network from the overparameterised students (EC, yellow stars), with few excess neurons.}
\end{figure}

\begin{figure}[!tbp]
  \centering
  \begin{minipage}[b]{0.495\textwidth}
    \includegraphics[width=\textwidth]{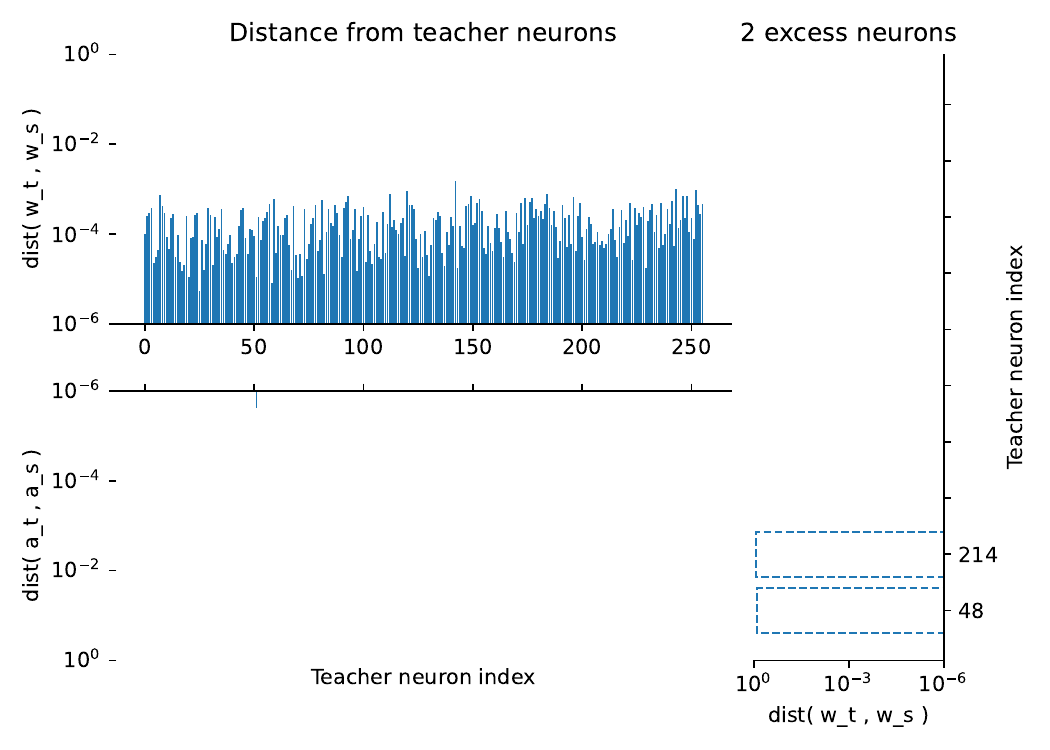}
    \caption{Reconstruction of $\sigma=\mathrm{g}$ network.}
  \end{minipage}
  \hfill
  \begin{minipage}[b]{0.495\textwidth}
    \includegraphics[width=\textwidth]{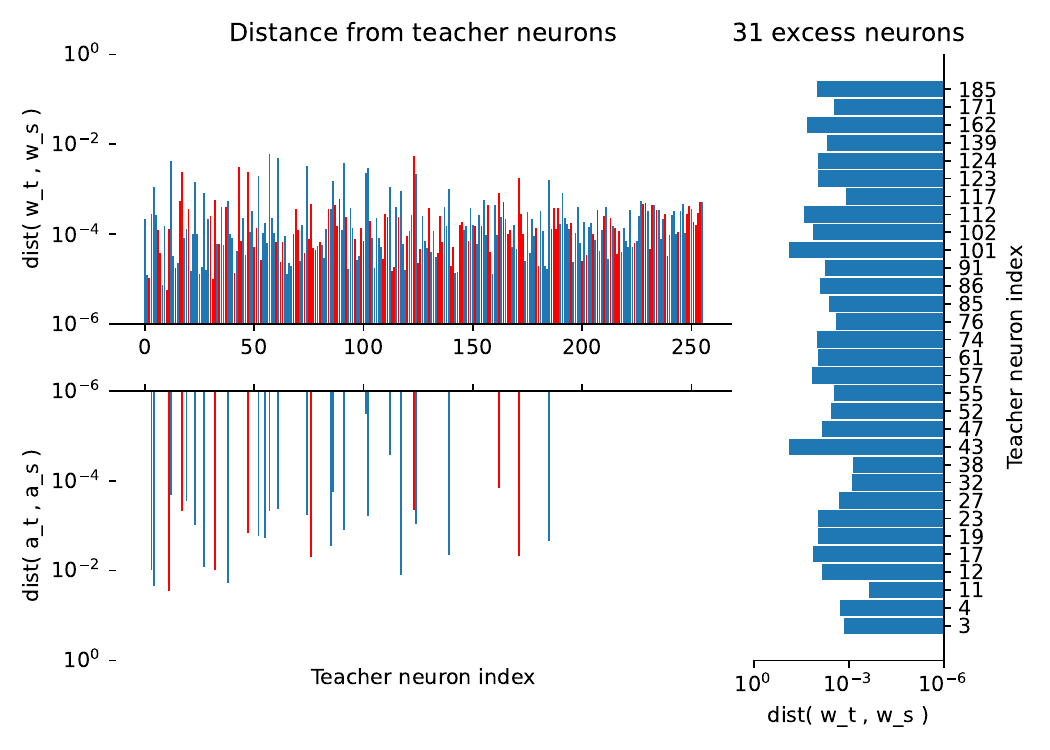}
    \caption{Reconstruction of $\sigma=\mathrm{sigmoid}$ network.}
  \end{minipage}
  \caption{{\bf Comparing reconstructed and target network of MNIST experiments in Table \ref{fig:6}A:} The left plots show the distance of each teacher neuron input (top) or output (bottom) weight vector with respect to the closest neuron of the reconstructed network. To indicate the unidentifiability of the sign of the weight vectors, we indicate in red the bars of neurons identified with the opposite sign. The right vertical plot shows how many excess neurons were found and to what teacher neuron they align the closest (in input weight vector); empty dashed bars indicate excess neurons with output weight vector norms below 0.1 (putatively zero neurons). We can still find duplicate neurons in reconstructed networks, they can be seen in these plots by having an excess neuron closely aligned to a teacher neuron and, consequently, a worse precision in alignment of the output weights (because of imprecisions in the sum of the duplicates' output weight vectors). The teacher neurons are ordered by descending output weight vector norm, as the lowest norm output weight neurons tend to be learnt at later stages of the learning process (i.e. at lower losses \cite{tian2020student}).}
\end{figure}

\begin{figure}[!tbp]
  \centering
  \begin{minipage}[b]{0.495\textwidth}
    \includegraphics[width=\textwidth]{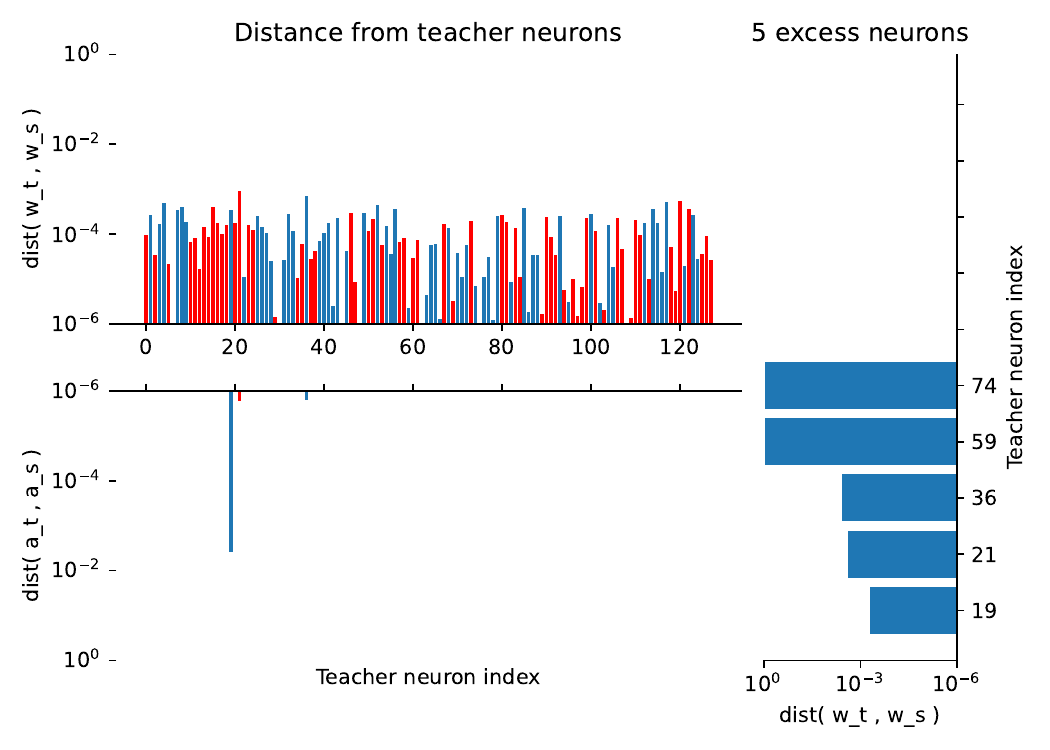}
    \caption{Reconstruction of $\sigma=\mathrm{tanh}$ network.}
  \end{minipage}
  \hfill
  \begin{minipage}[b]{0.495\textwidth}
    \includegraphics[width=\textwidth]{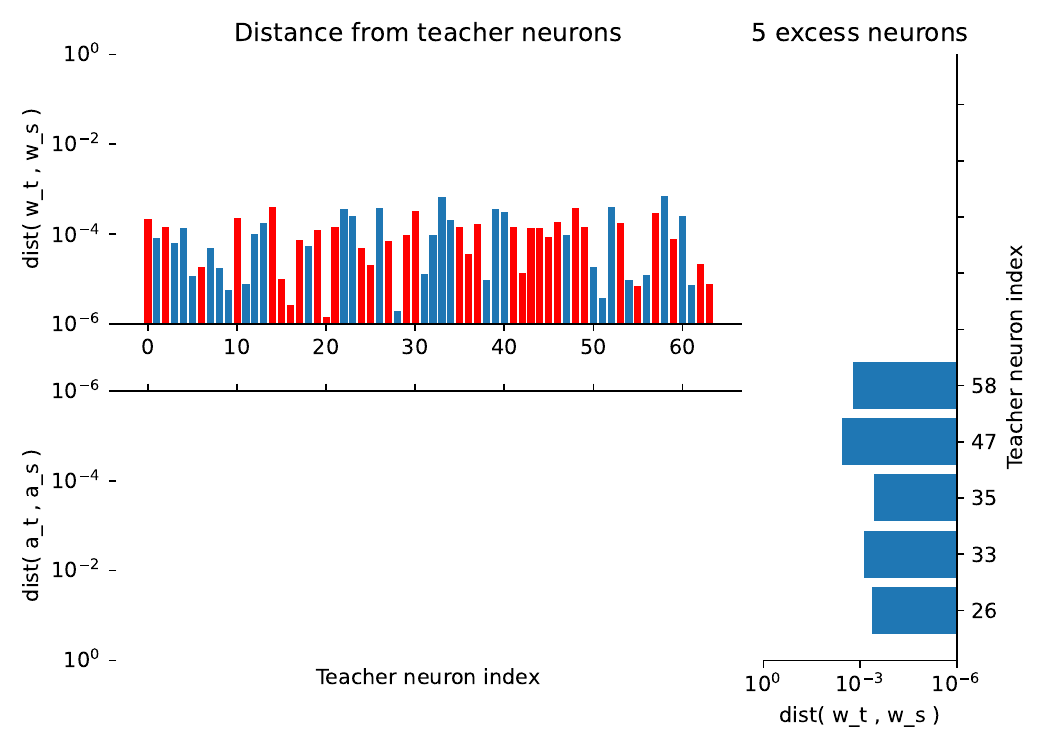}
    \caption{Reconstruction of $\sigma=\mathrm{softplus}$ network.}
  \end{minipage}
\end{figure}

\begin{figure}[!tbp]
  \centering
  \includegraphics[width=0.45\textwidth]{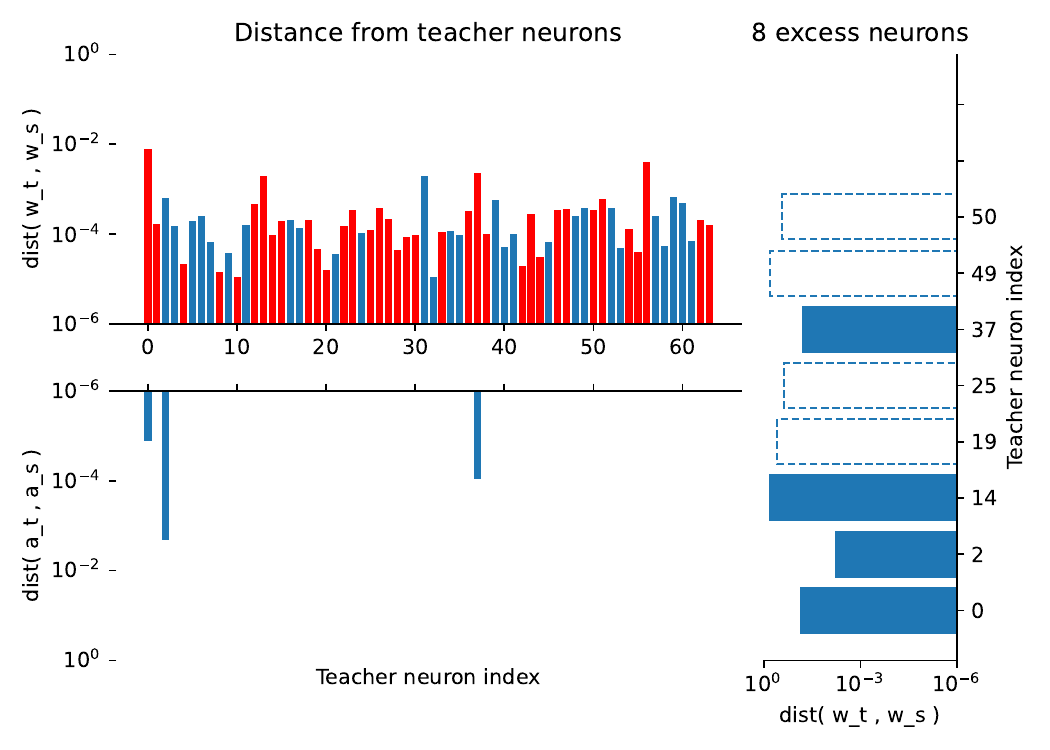}
  \caption{Reconstruction of $\sigma=\mathrm{relu}$ network.}
  \hfill
  \begin{minipage}[b]{0.495\textwidth}
  \end{minipage}
\end{figure}

\begin{figure*}[h]
   \begin{center}
   \centerline{\includegraphics[width=\textwidth]{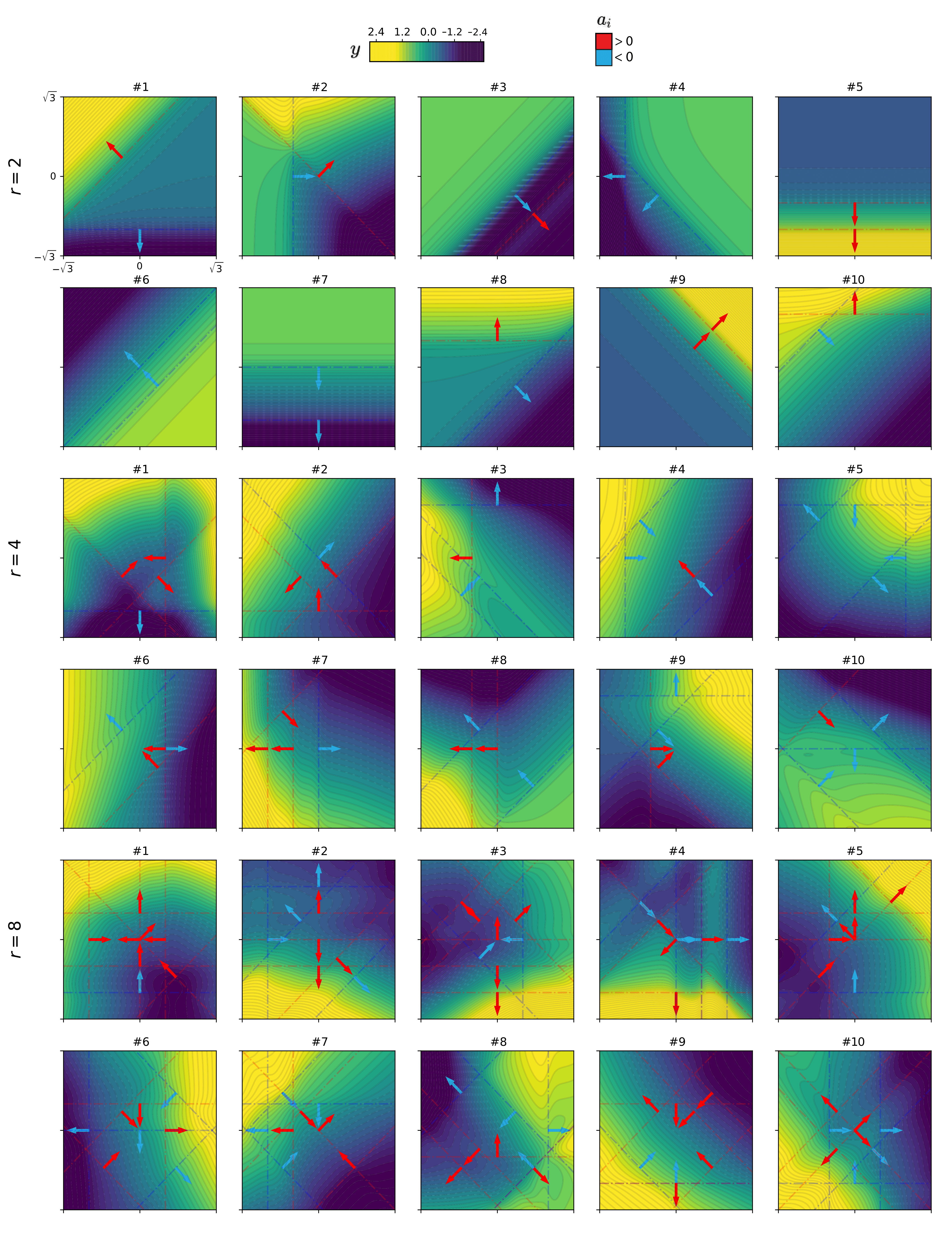}}
   \caption{{\bf Visualization of all $\mathbf{d_{in} = 2}$ teachers used for results in Figure \ref{fig:4} and \ref{fig:5}}}
   \label{fig:supp5}
   \end{center}
   \vskip -0.2in 
\end{figure*}

\clearpage

\end{document}